%% file: acl2025.tex
\pdfoutput=1
\documentclass[11pt]{article}

\usepackage[preprint]{acl}

\usepackage{times}
\usepackage{latexsym}
\usepackage{makecell}
\usepackage{hyperref}
\usepackage{tablefootnote}

\usepackage[T1]{fontenc}
\usepackage[utf8]{inputenc}

\usepackage{microtype}

\usepackage{inconsolata}

\usepackage{graphicx}
\usepackage{amsmath}
\usepackage{booktabs}

\newcommand{\factscore}{\textsc{FActScore}}
\newcommand{\SAFE}{\textsc{Safe}}

\newcommand{\BAFE}{\textsc{Bafe}}
\newcommand{\customlabel}[2]{%
\protected@write \@auxout {}{\string \newlabel {#1}{{#2}{}}}}

\definecolor{lightblue}{rgb}{0.68, 0.85, 0.9}
\definecolor{lightcoral}{rgb}{0.94, 0.5, 0.5}
\definecolor{melon}{rgb}{0.99, 0.74, 0.71}
\definecolor{forestgreen}{rgb}{0.13, 0.55, 0.13}

\title{How Does Response Length Affect Long-Form Factuality}

\author{James Xu Zhao,\quad Jimmy Z.J. Liu,\quad Bryan Hooi,\quad See-Kiong Ng\\
  National University of Singapore \\
  \texttt{xu.zhao@u.nus.edu} \\}

\begin{document}
\maketitle
\begin{abstract}

Large language models (LLMs) are widely used for long-form text generation. However, factual errors in the responses would undermine their reliability. Despite growing attention to LLM factuality, the effect of response length on factuality remains underexplored. In this work, we systematically investigate this relationship by first introducing an automatic and bi-level long-form factuality evaluation framework, which achieves high agreement with human annotations while being cost-effective. Using this framework, we conduct controlled experiments and find that longer responses exhibit lower factual precision, confirming the presence of length bias. To explain this phenomenon, we empirically examine three hypotheses: error propagation, long context, and facts exhaustion. Our results reveal that facts exhaustion, where the model gradually exhausts more reliable knowledge, is the primary cause of factual degradation, rather than the other two hypotheses.\footnote{Code and data are available at \url{https://github.com/XuZhao0/length-bias-factuality}.}

\end{abstract}

\section{Introduction}

Large language models (LLMs) are widely used for long-form text generation, such as long-form question answering~\citep{pmlr-v202-lee23n, xu-etal-2023-critical}, where responses may span hundreds or even thousands of words. However, these extended responses often contain factual errors, which are statements not grounded in established world knowledge~\citep{ji_survey_2023, 10.1145/3703155}. Such errors not only undermine the trustworthiness of LLMs, but also pose potential risks in high-stakes domains where factual accuracy is critical, such as healthcare~\citep{pal-etal-2023-med}.

\begin{figure}[!t]
    \centering
    \includegraphics[width=\linewidth]{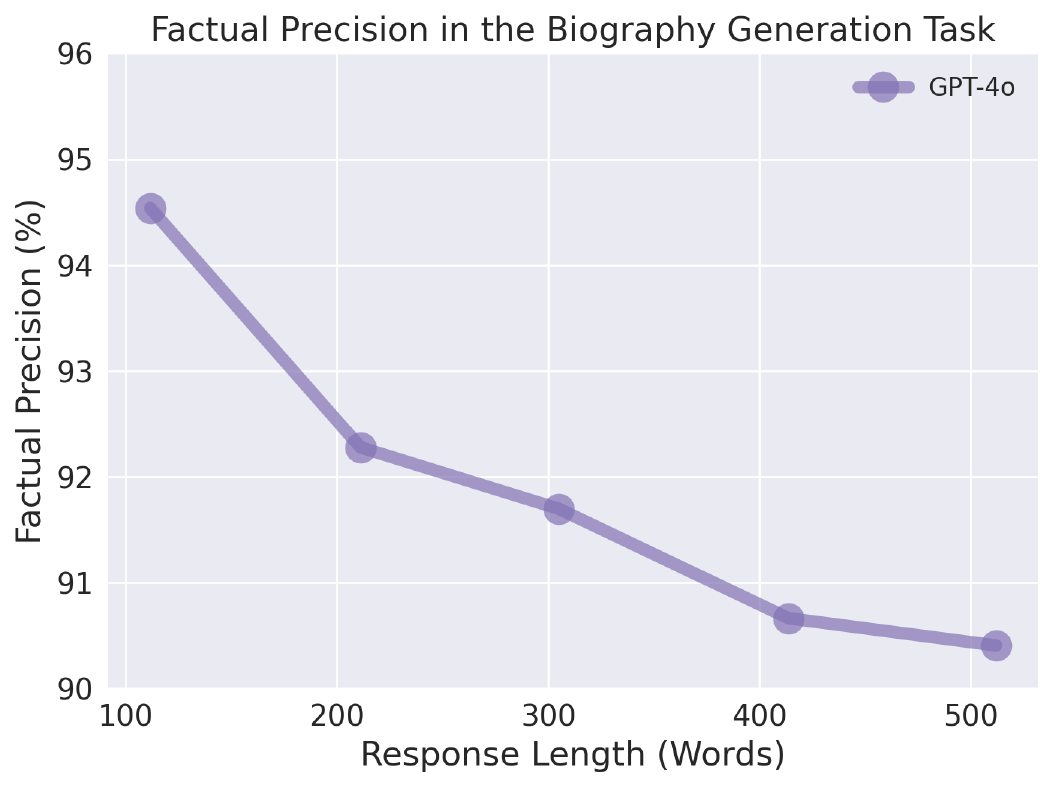}
    \caption{Factual precision gradually decreases as response length increases, which demonstrates the existence of length bias in long-form factuality. Response length is measured by word count (split by spaces).}
    \label{fig:length_bias}
\end{figure}

A key question in long-form text generation is whether and how response length affects factuality. Recent studies have reported different findings on this issue~\citep{NEURIPS2024_937ae0e8, zhou2024analyzing}, and there has been no systematic investigation into this problem. Without a clear understanding of how response length influences factuality, it remains challenging to develop effective strategies for reducing factual errors in long-form generation. To bridge this gap, our study focuses on the following research questions:

\begin{itemize}
    \item \textbf{RQ1}:Does response length affect factual precision? In other words, is there a length bias in long-form factuality?
    \item \textbf{RQ2}: If length bias exists, what are the underlying causes for it? 
\end{itemize}

To answer these research questions, there is a need for a reliable and efficient automatic long-form factuality evaluation framework. Existing evaluation methods, such as \factscore~\citep{min-etal-2023-factscore} and \SAFE~\citep{NEURIPS2024_937ae0e8}, have notable limitations (detailed in Section~\ref{sec:issues_factscore}). \factscore~verifies facts against a single retrieved Wikipedia page, which may miss factual information that is not present on the page. \SAFE~compares facts against Google Search results, but is highly time-consuming, requiring 28 minutes per response.

To address these limitations, we introduce \BAFE~(\textbf{B}i-level \textbf{A}tomic \textbf{F}act \textbf{E}valuation), an effective and efficient factuality evaluation method (Section~\ref{sec:bafe}). \BAFE~first decomposes long responses into atomic facts, each representing a piece of information. At the first level, each atomic fact is verified against a retrieved Wikipedia page. To reduce false negatives resulting from Wikipedia's limited coverage, unsupported facts go through the second-level verification using Google Search results. Through extensive human evaluation, \BAFE~achieves the highest agreement with annotators (89.31\%), while being 4 times faster and 7 times cheaper than \SAFE.

We then investigate \textbf{RQ1} (Section~\ref{sec:length-bias}) using \BAFE, to examine whether length bias exists in long-form factuality. We conduct experiments where we only vary the requested output length in the prompt to ensure a controlled analysis. We then measure the factual precision of the generated responses across different lengths. Our results (Figure~\ref{fig:length_bias},~\ref{fig:length_bias_longfact}) reveal a clear declining trend in factual precision as response length increases, providing empirical evidence that response length affects the factuality of LLM-generated text.

We further investigate \textbf{RQ2} (Section~\ref{sec:causes}) to identify the underlying causes of length bias in long-form factuality. We propose three possible explanations: (1) \textit{Error propagation}: Errors in earlier parts of the response propagate through the generation process, leading to an accumulation effect that results in factual degradation. (2) \textit{Long context}: Due to the autoregressive nature of LLMs, as the generation length increases, the model conditions on a longer content, making it more difficult to maintain factuality. (3) \textit{Facts exhaustion}: As response length increases, the model gradually exhausts more reliable knowledge it has, forcing it to rely on less certain or speculative details. 

We empirically validate these hypotheses through controlled experiments. Our results reveal that (1) \textit{Error propagation} exists but is weak and short-term, with no significant accumulation effect in longer responses. (2) \textit{Long context} does not degrade factual precision, as newly generated responses maintain consistent factual precision regardless of prior context length. (3) \textit{Facts exhaustion} is the primary cause of length bias. Specifically, we find that the model's factuality drops more when the model continuously generates on a single topic, as compared to when it generates on multiple topics. 

Our study provides a systematic analysis of the relationship between response length and factuality in long-form text generation. By confirming the presence of length bias, we demonstrate that longer responses inherently exhibit lower factual precision. Our findings on the underlying causes indicate that facts exhaustion, rather than error propagation or long context, is the primary reason for factual degradation. These insights contribute to a deeper understanding of length-related factuality challenges in LLMs and suggest potential directions for improving the factual accuracy in long-form generation.

\section{\BAFE: An Automatic and Bi-Level Long-Form Factuality Evaluator}
\label{sec:bafe}

\begin{figure*}[!ht]
    \centering
    \includegraphics[width=\linewidth]{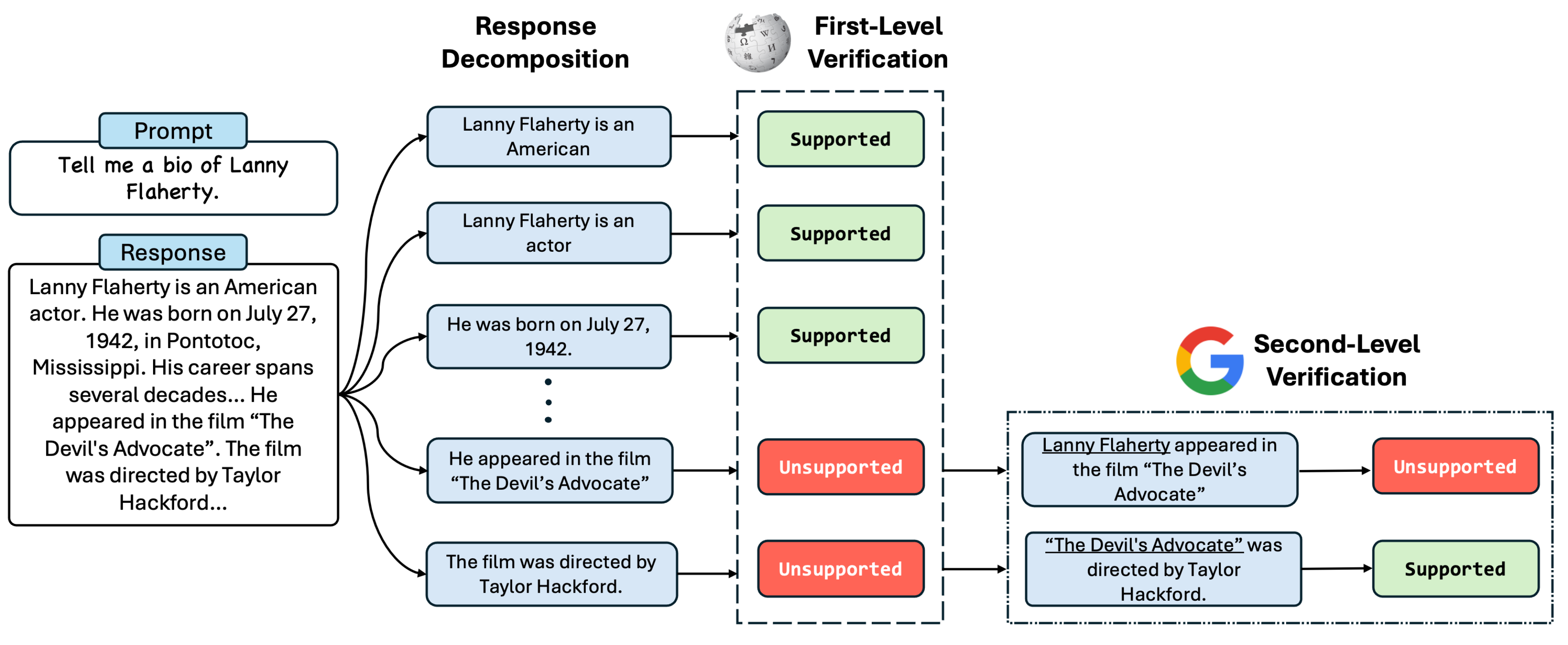}
    \caption{The pipeline of \BAFE, an automatic bi-level long-form factuality evaluator. It first decomposes long responses into atomic facts. Each atomic fact is compared with a retrieved Wikipedia page at first-level verification. Atomic facts that are not supported at first-level, will reach second-level verification, where they are revised to be self-contained and checked against Google Search results. Human evaluation shows superior performance of \BAFE~with a low cost in Section~\ref{sec:overview_BAFE}.}
    \vspace{-5pt}
    \label{fig:bafe_framework}
\end{figure*}

A reliable and efficient automatic long-form factuality evaluation framework is essential for investigating length bias in long-form text generation. In this section, we first analyze the limitations of existing methods (Section~\ref{sec:issues_factscore}). We then introduce \BAFE~(Bi-Level Atomic Fact Evaluation) in Section~\ref{sec:overview_BAFE}, and validate its effectiveness through comparison with human annotations (Section~\ref{sec:validate_bafe}).

\subsection{Issues with Existing Long-form Factuality Evaluation Method}
\label{sec:issues_factscore}

Two widely used methods for evaluating long-form factuality are \factscore~\citep{min-etal-2023-factscore} and \SAFE~\citep{NEURIPS2024_937ae0e8}. These approaches evaluate factual precision by decomposing responses into atomic facts and verifying them against Wikipedia or Google Search results.

\factscore~follows a two-step process: it first breaks a response into atomic facts, short statements that each contain one piece of information, and then validates each fact against retrieved Wikipedia pages. However, it has several limitations:

\begin{enumerate}
    \item \textbf{Limited information coverage}: This method relies exclusively on a single Wikipedia page for verification. It fails when relevant information is absent from the target page, but exists elsewhere. For example, when evaluating a generated biography of \textit{``Lanny Flaherty''}, the fact \textit{``The Devil’s Advocate was directed by Taylor Hackford''} is not supported by the Wikipedia page for \textit{``Lanny Flaherty''}, but is supported by the film's Wikipedia page. As a result, \factscore~incorrectly classifies supported facts as unsupported.
    \item \textbf{Reference ambiguity}: Since each atomic fact is evaluated individually, resolving ambiguous references becomes challenging. For example, the atomic fact stating \textit{``The film was directed by Taylor Hackford''} lacks sufficient context for verification. Without knowing which \textit{film} is being referenced, the evaluator cannot accurately assess factuality, leading to potential misclassification.
\end{enumerate}

\SAFE~improves upon \factscore. It agrees with the principle that long-form responses should be evaluated at the atomic fact level. Different from \factscore, \SAFE~verifies each atomic fact with Google Search results. \SAFE~also introduces two additional steps: (a) self-contained revision, which resolves reference ambiguity. (b) relevance checking, which filters out irrelevant atomic facts. Despite these improvements, \SAFE~has notable weaknesses:

\begin{enumerate}
    \item \textbf{Inefficiency and high cost}: Long-form responses usually contain hundreds of atomic facts, each requiring multiple processing steps: (1) self-contained revision, (2) relevance check, (3) search query generation, (4) Google Search execution, and (5) final factuality determination. Since query generation and search steps are repeated multiple times (5 times by default), this process imposes substantial computational overhead. It takes 28 minutes to evaluate a response without parallelization and costs \$0.5, making it computationally expensive and less accessible.

    \item \textbf{Unnecessary relevance filtering}: The relevance checking step, designed to exclude irrelevant claims, introduces unnecessary complexity. As current large language models have demonstrated strong instruction following capabilities~\citep{NEURIPS2022_b1efde53}, we do not observe completely irrelevant statements in generated responses. Prior research~\citep{song-etal-2024-veriscore} has shown that this filtering step negatively impacts the evaluation, as it incorrectly removes relevant facts. Moreover, this step increases processing time and cost.
\end{enumerate}

\subsection{\BAFE~: Bi-Level Atomic Fact Evaluation}
\label{sec:overview_BAFE}
To address the above limitations of existing factuality evaluation methods, we propose \BAFE~(\textbf{B}i-Level \textbf{A}tomic \textbf{F}act \textbf{E}valuation), as illustrated in Figure~\ref{fig:bafe_framework}. The evaluation process consists of three steps:

\begin{enumerate}
    \item \textbf{Response decomposition}: The long-form response is first decomposed into a series of atomic facts, each representing a factual statement.
    \item \textbf{First-level verification}: Each atomic fact is compared against a retrieved Wikipedia page. We use an LLM to judge whether the fact is supported.
    \item \textbf{Second-level verification}:  If an atomic fact is unsupported at the first level, it is revised to be self-contained. We then issue a single Google Search query and compare the fact with search results, using a similar LLM-based judge as in the first level. This level complements first-level verification by allowing for broader information coverage. If a fact remains unsupported after both levels, we consider it to be a factual error.
\end{enumerate}

The bi-level verification framework improves both accuracy and efficiency by: (1) expanding knowledge coverage, leveraging Wikipedia as a reliable knowledge source and performing Google Search for broader fact-checking; and (2) enhancing efficiency. Only Wikipedia-unsupported facts undergo second-level verification. We remove the unnecessary relevance filtering step to reduce computational overhead. Additionally, we only issue a single Google Search query per fact, as we observe that multiple queries do not consistently yield more relevant information. 

We provide details on the implementation of \BAFE~in Appendix~\ref{appendix:implementation_detail_bafe}.

\subsection{Validate the Effectiveness of \BAFE}
\label{sec:validate_bafe}
To validate the effectiveness of \BAFE, we conduct human evaluation on 786 atomic facts from the \textit{biography generation task}~\citep{min-etal-2023-factscore}. Unlike \factscore, where annotators are restricted to preset Wikipedia pages, our human evaluators have full access to the Internet. Three annotators independently evaluate each fact, achieving a Fleiss $\kappa$ score of 0.7655 (substantial agreement)~\citep{fleiss1971measuring}. The majority vote among annotators serves as the ground truth. Details of the annotation process are provided in Appendix~\ref{appendix:human_eval}. 

\input{tables/evaluation_method_comparison}

The results in Table~\ref{tab:method_comparison} demonstrate that \BAFE~achieves the highest agreement with human annotations, reaching 89.31\%. In comparison, \factscore~achieves only 69.97\% agreement, primarily due to its limited information coverage, as discussed in Section~\ref{sec:issues_factscore}. \SAFE~achieves a higher agreement rate of 84.48\%, but at a substantially higher computational cost, requiring 28 minutes and \$0.49 per response. In contrast, our method is over 7 times cheaper and 4 times faster than \SAFE.

Notably, \BAFE~outperforms \SAFE~despite issuing only a single search query per atomic fact, whereas \SAFE~performs five queries per fact. We attribute this to two factors: (1) Wikipedia provides more holistic information, whereas Google Search returns web snippets that often lack sufficient context and may introduce misleading information. This supports our design choice of using Wikipedia as the primary verification source, with Google Search serving as a complementary fallback. (2) More searches do not necessarily yield more valid information. In practice, we observe many duplicate search results in \SAFE, leading to redundancy. Moreover, excessive search results may introduce noise, distracting the evaluator and increasing the likelihood of misjudgment. Further analysis and case studies are provided in Appendix~\ref{appendix:case_comparison_safe}.

By achieving high accuracy while maintaining efficiency, \BAFE~provides a robust foundation for large scale experimental analysis, enabling a systematic investigation into the relationship between response length and long-form factuality.

\section{Is There a Length Bias in Long-Form Factuality?}
\label{sec:length-bias}

In this section, we examine RQ1, whether length bias exists in long-form factuality, using \BAFE. We conduct controlled experiments by only varying the requested response length and analyze how factual precision changes accordingly.

\subsection{Experimental Setup}
\paragraph{Datasets.} To investigate length bias in LLM-generated text, we conduct experiments on two datasets, the \textit{biography generation task} and the \textit{long fact description task}. We select these tasks because the generated responses typically contain specific and verifiable statements, rather than subjective or debatable claims, making it well-suited for factuality analysis.

The \textit{Biography generation task}~\citep{min-etal-2023-factscore} includes 183 people's names, covering diverse professions and different levels of rarity. 

The \textit{Long fact description task} involves generating long text that describes entities from different categories. We select 140 non-person entities, such as ``Hendra virus'', from LongFact-Concepts~\cite{wang-etal-2024-factuality}. This dataset spans 4 broad categories: humanities, STEM, social science, and others, further subdivided into 26 topics, such as music and chemistry. Additional details of this dataset are provided in Appendix~\ref{appendix:long_fact_dataset}.

\paragraph{Model and prompt setting.} We conduct all experiments using GPT-4o\footnote{GPT-4o version: \textit{gpt-4o-2024-08-06}.}. We use consistent prompt formats: \textit{``Tell me a bio of <entity>''} for the biography generation task, and \textit{``Tell me about <entity>''} for the long fact description task. To control the response length, we append an instruction in the system prompt: \textit{``Generate with around <x> words''}, where $x \in \{100, 200, 300, 400, 500\} $. This setting leverages GPT-4o's strong instruction-following capability~\citep{openai2024gpt4ocard}, allowing us to control response length while maintaining consistency across prompts.\footnote{We use greedy decoding for all our experiments.}

\paragraph{Evaluation.} We evaluate the factuality of long-form responses using \BAFE. Following~\citet{min-etal-2023-factscore}, we use \textit{factual precision} as the evaluation metric, defined as the percentage of supported facts among all atomic facts in the response.

\begin{figure}[!t]
    \centering
    \includegraphics[width=\linewidth]{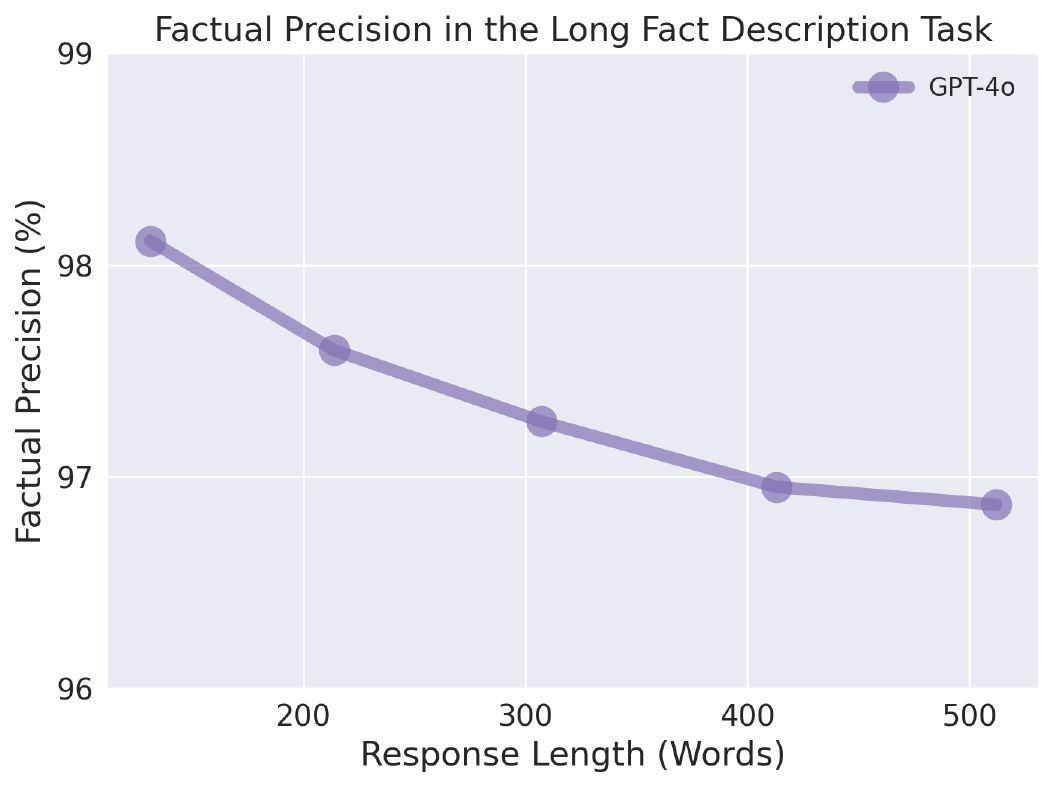}
    \caption{Factual precision decreases as response length increases in the long fact description task. Response length is measured by word count, using space as the delimiter.}
    \label{fig:length_bias_longfact}
\end{figure}

\subsection{Results and Discussion}
\paragraph{Length bias exists in long-form factuality.} Experiment results in Figure~\ref{fig:length_bias} and Figure~\ref{fig:length_bias_longfact} demonstrate that as the response length increases, the factual precision gradually decreases. In the biography generation task, when the model is instructed to generate a 100-word response, the factual precision is around 94.5\%. However, as the response length increases to 200 words, the factual precision drops to 92.2\%, with a reduction of 2.3\%. This downward trend continues, with the factual precision further decreasing to 90.5\% for responses with around 500 words. A similar degradation trend is observed in the long fact description task. The factual precision decreases from 98.1\% to 96.9\%, as the response length increases from 100 to 500 words. 

These results provide strong evidence that longer responses are prone to have lower factual precision, confirming the presence of length bias in long-form factuality. This observation motivates us to investigate RQ2: What causes length bias in LLM-generated responses? 

\section{What Causes Length Bias in Long-Form Factuality?}
\label{sec:causes}

To understand the underlying causes for length bias in long-form factuality (RQ2), we investigate potential reasons in this section. We first propose the following hypotheses:

\begin{enumerate}
    \item \textbf{Error propagation}: Errors in one part of the response can propagate to subsequent parts. This accumulation of errors leads to a decline in factual precision in longer responses~\citep{min-etal-2023-factscore}.
    \item \textbf{Long context}: As the generation length increases, LLMs must condition on a longer sequence of preceding tokens, which makes it more difficult to maintain coherence and factual accuracy, thereby increasing the likelihood of factual errors.
    \item \textbf{Facts exhaustion}: The model prioritizes the more reliable knowledge it has, when generating short responses, avoiding speculative or less supported claims. However, as the response length increases, the model is forced to include additional details, some of which it is less confident about, leading to more hallucinated content.
\end{enumerate}

In the following subsections, we empirically validate these hypotheses through controlled experiments in the \textit{biography generation task}.

\subsection{Effect of Error Propagation}
\label{sec:effect_error_propation}
In this subsection, we investigate the effect of error propagation with autocorrelation analysis and counterfactual analysis.

\subsubsection{Autocorrelation Analysis}
\label{sec:autocorrelation_analysis}
Autocorrelation analysis~\citep{therrien2018probability} is a statistical method commonly used to measure the relationship between a variable and its lagged values. In our study, we construct a binary error series for each response, by assigning 0 to supported atomic facts, and 1 to unsupported atomic facts. We compute the autocorrelation coefficient $r_k$ at different lags ($k=0, 1, ..., 8$) using the Equation~\ref{eq:autocorrelation}:

\begin{equation}
    r_k = \frac{\sum_{t=1}^{N-k} (x_t - \bar{x})(x_{t+k} - \bar{x})}{\sum_{t=1}^N (x_t - \bar{x})^2}
    \label{eq:autocorrelation}
\end{equation}

where $x_t$ is the binary error value at position $t$. $\bar{x}$ is the mean of the binary error series. $N$ is the length of the binary error series. $k$ is the lag, the number of positions offset between the compared values. 

\begin{figure}[!t]
    \centering
    \includegraphics[width=\linewidth]{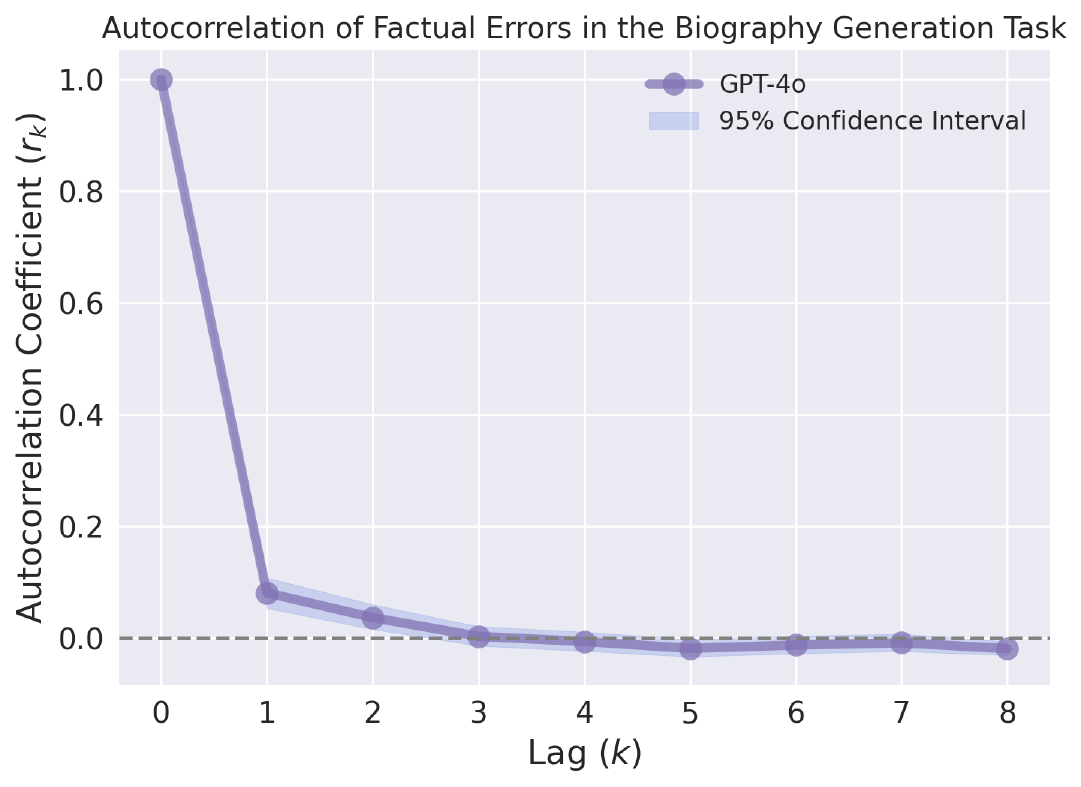}
    \caption{Autocorrelation coefficient at different lags. Results are aggregated over all responses to compute the average autocorrelation coefficient for each lag. The 95\% confidence intervals are obtained with 2000 times bootstrap resampling. Only the coefficient at lag 1 is slightly higher than 0 with statistical significance.}
    \label{fig:autocorrelation_analysis}
\end{figure}

\paragraph{Error propagation has only a minor short-term effect.} As shown in Figure~\ref{fig:autocorrelation_analysis}, the autocorrelation coefficient at lag 1 is positive and statistically significant, suggesting that if an unsupported fact appears, there is an increased likelihood that the next fact will also be unsupported. However, its small magnitude, around 0.1, suggests that the effect is weak. From lag 1 onward, the autocorrelation coefficients approach zero, indicating no statistically significant long-term dependency between errors. This suggests that errors are only weakly correlated and do not accumulate over the entire response.

\subsubsection{Counterfactual Analysis}
\label{sec:counterfactual_analysis}
To complement the statistical findings, we perform a counterfactual analysis to examine the causal impact of an early factual error on subsequent factual precision. 

We start by generating a biography in the same way as in our previous experiments. From the output, we split the first sentence from the biography. For example, \textit{``Harrison Ford is an American actor, born on \textbf{July 13, 1942}''}. We then create a counterfactual variant by flipping the factual correctness: \textit{``Harrison Ford is an American actor, born on \textbf{July 14, 1943}''}. We use both the original and flipped versions as the starting point to generate the rest of the biography. In this way, we have two biographies with different settings:

\begin{itemize}
    \item \textbf{Vanilla setting}: The generation continues based on the original first sentence.
    \item \textbf{Flipped factuality setting:} The generation continues based on the counterfactual first sentence.
\end{itemize}

We compute the factual precision for three segments: (1) the first sentence, (2) the second sentence, and (3) all sentences following the first in the biography. Results are shown in Table~\ref{tab:counterfactual_analysis_results}.

\input{tables/counterfactual_analysis_results}

\paragraph{Factual errors do not propagate throughout the response.} The flipped first sentence has 14.13\% lower factual precision than the original one, as expected. However, the factual precision of the second sentence decreases by only 0.5\%. More importantly, the factual precision of all subsequent sentences after the first is even slightly higher in the flipped setting compared to the vanilla setting. One possible explanation is that flipping the factual correctness of the first sentence alters the response structure. For example, the average number of facts in the subsequent sentences decreases from 60.4 (vanilla setting) to 58.6 (flipped factuality setting), with some details omitted. An illustrative example is provided in Appendix~\ref{appendix:counterfactual_examples}.

\paragraph{Conclusion on error propagation.} Both analyses consistently indicate that error propagation has limited short-term effects and is not the main cause of factual degradation as response length increases. 

\subsection{Effect of Long Context}
\label{sec:effect_long_context}

To investigate the effect of long context, we increase the context length while evaluating the factual precision of newly generated content. We prompt LLMs to generate responses of two sequential sections on different topics, so that observed variations in factual precision are attributable to the preceding context length, rather than other factors. Specifically, the responses are structured in the following manner:

\begin{itemize}
    \item \textbf{Context section (Topic A)}: We vary the biography length in this section to control the preceding context length.
    \item \textbf{Evaluation section (Topic B)}: We fix the biography length in this section and limit our evaluation to this section only.
\end{itemize}

We experiment with two different topics for \textit{Topic A}: \textit{``Early life''} and \textit{``Personal life''}, with context lengths ranging from 100 to 500 words. For \textit{Topic B}, we consistently use \textit{``Career''} with a fixed length of 200 words.
 
\begin{figure}[!t]
    \centering
    \includegraphics[width=\linewidth]{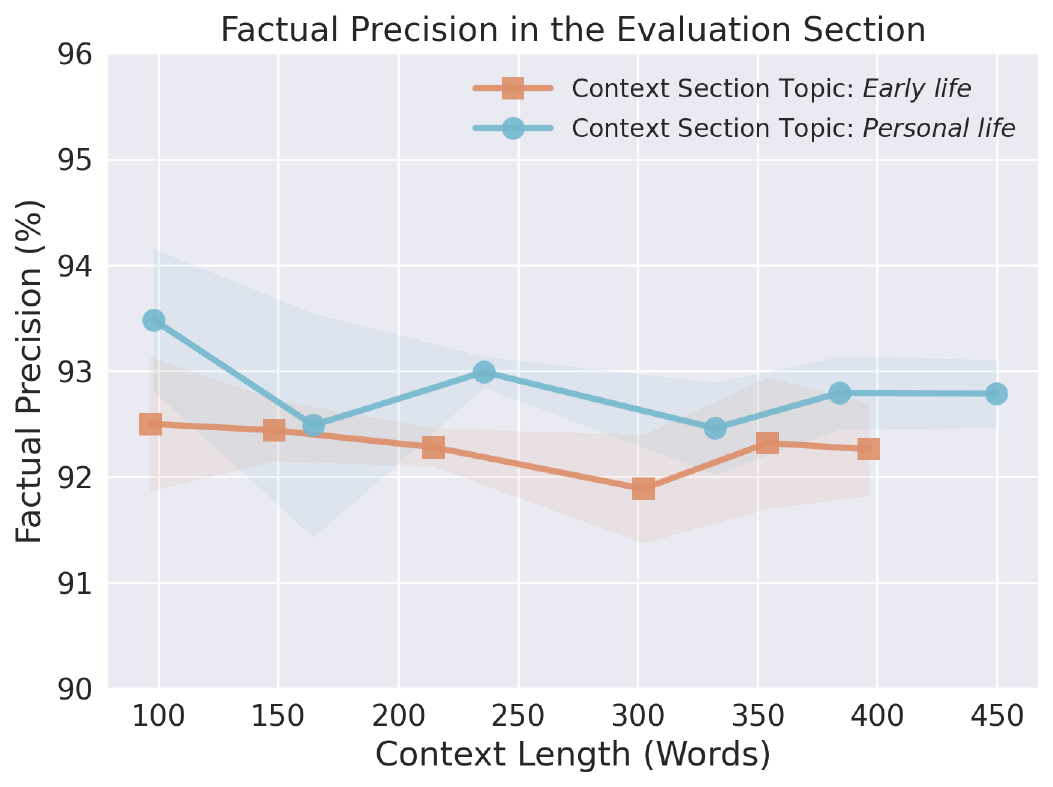}
    \caption{Factual precision in the evaluation section across varying context lengths and topics. The evaluation section is set to \textit{``Career''} topic with fixed length. Results are obtained over three runs. As the context length increases, factual precision does not decline significantly.}
    \label{fig:context_length}
\end{figure}

\paragraph{Long context is not the cause of factual degradation.} Experimental results in Figure~\ref{fig:context_length} show that factual precision for \textit{Topic B} remains stable across all variations in context length and \textit{Topic A} settings. Specifically, when \textit{Topic A} is set to \textit{``Early life''}, the factual precision of \textit{``Career''} responses remains around 92.5\%. As the context length increases from 100 words to around 400 words, there is a minimal decline for factual precision from 92.50\% to 92.26\%, with a negligible 0.24\% difference. Similarly, when \textit{Topic A} is set to \textit{``Personal Life''}, factual precision remains almost unaffected. At a context length of around 150 words, the factual precision is 92.49\%, and it even slightly increases to 92.79\% when the context length reaches 450 words.

These results indicate that increasing context length does not significantly affect the model's factual precision. Therefore, long context is not a primary cause of factual degradation in long-form text generation. This motivates further investigation into other explanations, such as facts exhaustion.

\subsection{Effect of Facts Exhaustion}
\label{sec:effect_fact_exhaustion}
\begin{figure}[!t]
    \centering
    \includegraphics[width=\columnwidth]{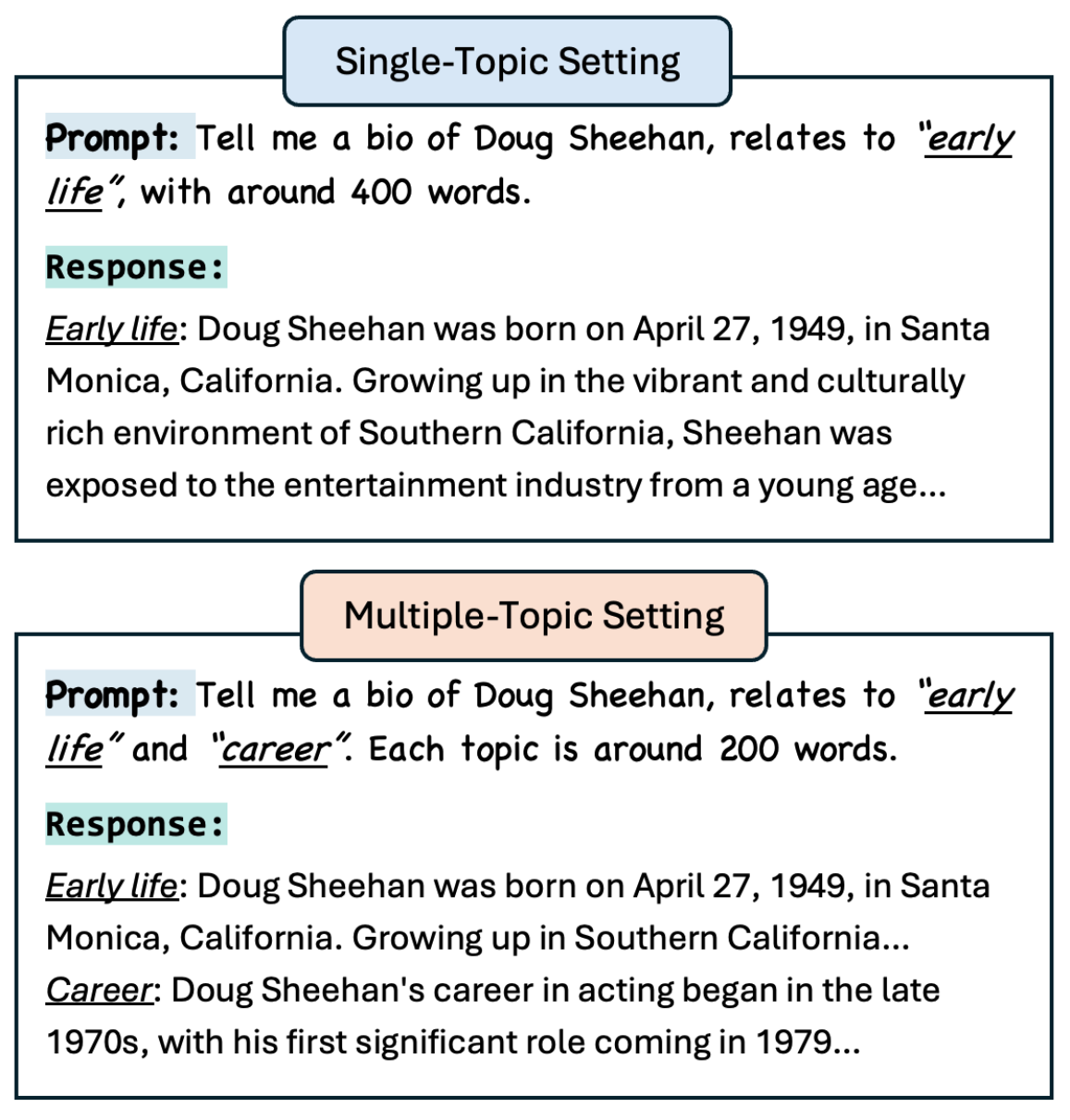}
    \caption{An example of facts exhaustion experiment setting. (1) Single-Topic Setting: The model generates a 400-word response focused on a single topic. (2) Multiple-Topic Setting: The model generates a 400-word response covering two topics, with 200 words per topic. Full prompts are provided in Appendix~\ref{appendix:example_prompts}.}
    \label{fig:exhaustion_setting_example}
\end{figure}

To validate the facts exhaustion hypothesis, we examine whether forcing the model to generate long responses on a single topic results in lower factual precision, compared to allowing the model to cover multiple topics. We consider two experimental settings, \textbf{single-topic setting} and \textbf{multiple-topic setting}, as illustrated in Figure~\ref{fig:exhaustion_setting_example}.

To ensure a fair comparison, we aggregate facts across prompts so that both settings cover the same two topics and each topic contributes an equal number of words. For example, when using the topic pair \textit{``Early life''} and \textit{``Career''}, we set up:

\begin{itemize}
    \item \textbf{Single-Topic Setting}: (1) Tell me a bio of <entity> related to ``early life'' with around 400 words. (2) Tell me a bio of <entity> related to ``career'' with around 400 words
    \item \textbf{Multiple-Topic Setting}: (1) Tell me a bio of <entity> related to ``early life'' and ``career'', each with around 200 words. (2) Tell me a bio of <entity> related to ``career'' and ``early life'', each with around 200 words.
\end{itemize}

We then aggregate the facts from the two prompts in each setting, which allows us to isolate the effect of topic selection and content quantity. In our experiments, topics are chosen from a fixed set of three commonly used biography sections: \textit{``Early life''}, \textit{``Personal life''} and \textit{``Career''}.

\begin{figure}[!t]
    \centering
    \includegraphics[width=\linewidth]{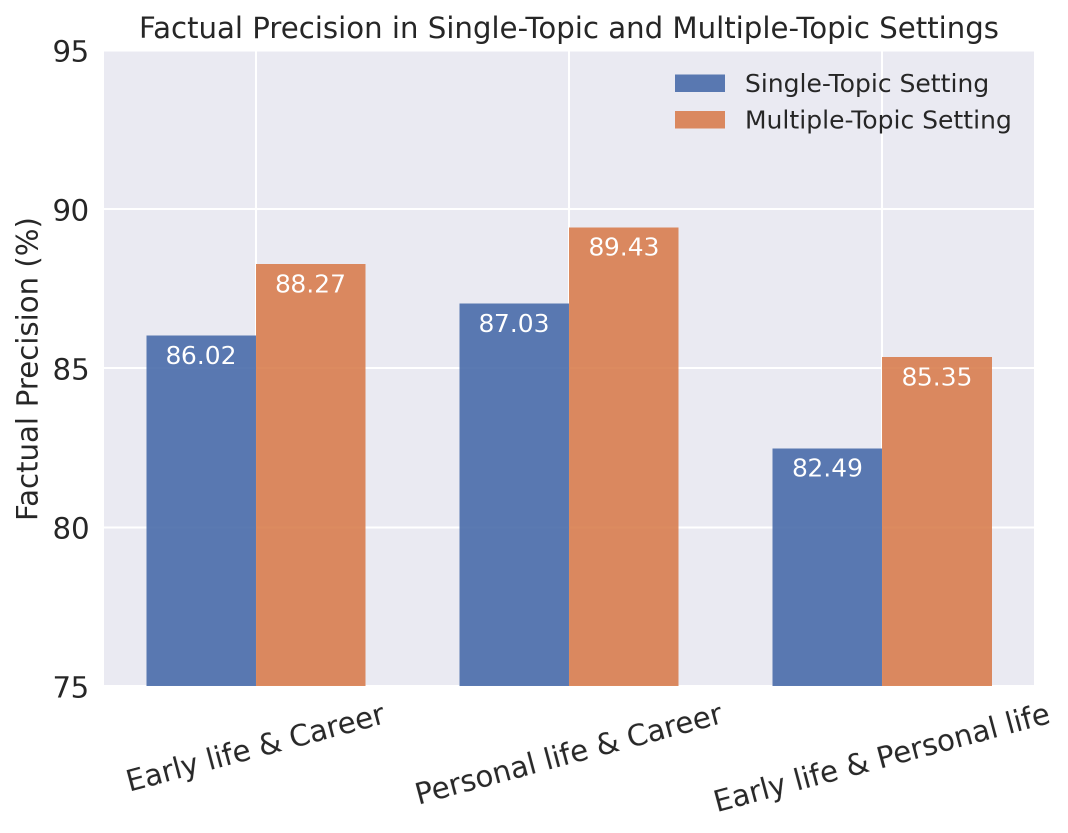}
    \caption{Factual precision in single-topic and multiple-topic settings across different topic pairs. Responses in multiple-topic settings consistently have higher factual precision than those in single-topic settings.}
    \label{fig:exhaustion_results}
\end{figure}

\paragraph{Multiple-topic setting consistently leads to higher factual precision.} Results in Figure~\ref{fig:exhaustion_results} demonstrate that in all cases, responses generated in the multiple-topic setting have higher factual precision compared to those in the single-topic setting, with improvements ranging from 2.25\% to 2.86\%. For instance, for the topic pair \textit{``Early life''} and \textit{``Career''}, factual precision improves from 86.02\% to 88.27\%. Notably, for \textit{``Early Life''} and \textit{``Personal Life''}, factual precision in the single-topic setting is 82.49\%, whereas allowing the model to switch topics in the multiple-topic setting increases the precision to 85.35\%. 

\paragraph{Qualitative analysis further supports the facts exhaustion hypothesis.} To complement the quantitative findings, we examine model responses of varying lengths and observe a consistent trend: longer responses tend to include more speculative or filler content. As illustrated in Table~\ref{tab:example_response_length}, the 100-word response includes the statement \textit{``later became the acting Chief Justice''}, while the 200-word response adds \textit{``In 2009, he was elevated to Chief Justice''}, which is factually incorrect. Such added details often lack factual grounding and are more likely to be unsupported.

These results suggest that when the model is forced to generate long responses, it gradually exhausts reliable factual knowledge and begins compensating by including speculative or inaccurate details, leading to a decline in factual precision. This supports the facts exhaustion hypothesis, where longer responses demand more uncertain information, reducing factual accuracy.

\paragraph{Conclusion on RQ2.} Taken together, our study provides strong empirical evidence that facts exhaustion, rather than error propagation or long context, is the primary cause of factual precision degradation in long-form responses.

\begin{table*}[t]
    \scriptsize
    \begin{tabular}{c|p{13cm}}
    \toprule
    \makecell[c]{\textbf{Requested Output} \\\textbf{Length (Words)}} & \makecell[c]{\textbf{Responses}} \\
     \midrule
     100 & {\raggedright
            Nisar Ahmad Kakru is a distinguished Indian jurist who has made significant contributions to the legal field in India. \textcolor{red}{Born on October 25, 1949}, in the Baramulla district of Jammu and Kashmir, he pursued his education in law and embarked on a career that would see him rise to prominent judicial positions. Kakru served as the Chief Justice of the High Court of Andhra Pradesh from February 2010 to October 2011. Prior to this, he was a judge at the Jammu and Kashmir High Court, where he was appointed in 1997 and \textit{later became the acting Chief Justice}. His tenure is noted for his commitment to justice and legal reform.
        }  \\
    \midrule
     200 & {\raggedright
             Nisar Ahmad Kakru is a distinguished Indian jurist known for his significant contributions to the judiciary in India. \textcolor{red}{Born on October 25, 1949}, in the region of Jammu and Kashmir, Kakru pursued a career in law, which led him to become a prominent figure in the Indian legal system. He completed his law degree and began his legal practice, eventually gaining recognition for his expertise and dedication. Kakru's judicial career saw a steady rise as he was appointed as a judge of the Jammu and Kashmir High Court in 1997. \textcolor{red}{His tenure was marked by a commitment to justice and fairness earning him respect among his peers} and the legal community. \textcolor{red}{\textit{In 2009, he was elevated to the position of Chief Justice} of the Andhra Pradesh High Court}...\textit{(Truncated due to space limitation)}
     } \\
     \bottomrule
    \end{tabular}%
    \caption{Examples of responses with different lengths (100 and 200 words). Unsupported statements are highlighted with \textcolor{red}{red} color. Compared to a 100-word response, the 200-word response includes more details, such as\textit{ ``In 2009, he was elevated...''}, which are unsupported. More examples are provided in Appendix~\ref{appendix:example_response_length}.}
    \label{tab:example_response_length}
\end{table*}

\section{Related Work}

\subsection{Factuality of LLMs}
Factuality refers to the degree to which information is accurate and grounded in established knowledge~\citep{10.1145/3571730, zhang2023siren, NEURIPS2024_937ae0e8}. It has been a long-standing issue in large language models (LLMs)~\citep{maynez-etal-2020-faithfulness, chen-etal-2023-beyond, wang-etal-2024-factuality, Augenstein2024FactualityCI}. Prior research has identified multiple factors contributing to factual errors. \citet{zheng2024why} attribute factual errors to knowledge gaps. \citet{zhang2024how} find that propagated errors are recognized by the language model itself. But both studies evaluate short-form question-answering tasks. \citet{orgad2025llms} explore the internal encoding of LLMs, finding that in long-form generation, truthfulness information is localized within the specific answer token. Our work provides an empirical investigation of factuality in long-form responses, systematically analyzing how response length affects factuality.

\subsection{Length Bias}
The presence of length bias in LLMs has been widely discussed. \citet{NEURIPS2023_91f18a12} observe that when LLMs serve as a judge, they favor longer responses. \citet{liu-etal-2024-lost} find that model performance degrades when accessing information in the middle of a long input context. Many studies have focused on long-context understanding, evaluating how models process and utilize extended input text~\citep{shaham-etal-2023-zeroscrolls, dong-etal-2024-bamboo, bai-etal-2024-longbench, NEURIPS2024_c0d62e70, jacovi2025factsgroundingleaderboardbenchmarking}. However, these works primarily explore on input length, rather than generation length. In the context of long-form generation, \citet{min-etal-2023-factscore} and \citet{tu2024investigatingfactualitylongformtext} find that later-generated text contains more factual errors. \citet{NEURIPS2024_937ae0e8} observe a decline in factual precision over longer outputs, though their evaluation lacks systematic investigation. \citet{zhou2024analyzing} find that longer descriptions do not lead to more hallucinations. Compared to prior work, we investigate the effect of length bias on factuality in long-form text generation, and further identify its underlying causes.

\subsection{Automatic Long-Form Factuality Evaluation}
Automatically evaluating factuality in long-form text generation is challenging, as long-form responses often contain a mixture of correct and incorrect information. Prior approaches attempt to address this issue through response decomposition and atomic fact verification. \factscore~\citep{min-etal-2023-factscore} and \SAFE~\citep{NEURIPS2024_937ae0e8} are two widely used methods. But both methods have notable limitations as detailed in Section~\ref{sec:issues_factscore}. \citet{song-etal-2024-veriscore}~propose \textsc{VeriScore} to evaluate verifiable claims only. \citet{lin-etal-2024-advancing}~propose D-FActScore to evaluate factuality in content with ambiguous entities. Compared to prior long-form factuality evaluation works, \BAFE~introduces a bi-level verification framework, incorporating both retrieved Wikipedia pages and broad Google Search results to improve factuality evaluation accuracy.

\section{Conclusion and Future Work}
In this work, we systematically investigate how response length affects long-form factuality. We first propose \BAFE, an automatic and bi-level factuality evaluation framework, to facilitate our investigation. Using \BAFE, we conduct extensive experiments and confirm the existence of length bias, where longer responses exhibit lower factual precision. Our empirical findings reveal that \textit{facts exhaustion}, rather than error propagation or long context, is the primary cause of factual degradation in long-form generation.

Our findings suggest several directions for future work: (1) Refining long-form factuality evaluation metrics. Due to the existence of length bias, we encourage developing more comprehensive metrics that consider both factual coverage and precision. (2) Mitigating facts exhaustion. As \textit{facts exhaustion} is the primary cause of factual degradation, rather than others. Future work could explore methods to supplement or retrieve deeper factual knowledge in LLMs to improve long-form factual accuracy.

\section*{Limitation}
First, our evaluation method, \BAFE, is designed for fact-intensive tasks, where all statements are assumed to be verifiable. However, in some cases, statements may be ambiguous or unverifiable, posing challenges for factuality evaluation. Moreover, while \BAFE~incorporates multiple knowledge sources, it may still lack coverage of specialized knowledge that requires domain-specific references, such as scientific literature. Expanding the framework to include additional retrieval sources, such as journal articles, could further improve the reliability of the evaluation.

Second, our experiments are primarily conducted on GPT-4o, given its strong instruction-following capability and widespread use. Some of our experimental designs rely on response length control, which may not generalize well to LLMs with weaker instruction-following capability. Future work should explore whether our findings hold across a broader range of models, such as open-source LLMs.

Third, due to the black-box nature of LLMs, it is hard to directly examine the \textit{``facts exhaustion''} problem at the internal knowledge level. So we choose to approach this problem from an empirical observation perspective. Future work could explore strategies to better understand the internal knowledge usage and depletion in LLMs. 

\section*{Ethic Statements}
Our research focuses on factuality in long-form text generation. The potential ethical impact is the implications of factual errors in LLM-generated content. As part of our study, we conducted a human evaluation. Each annotator signed a consent form, explicitly agreeing that their annotations may be used for scientific research and publication. No personally identifiable information was collected. Each annotator was paid \$15 per hour. The human evaluation protocol was reviewed and approved by NUS Department Ethics Review Committee (DERC).

\section*{Acknowledgment}
We thank Prof. Hwee Tou Ng and Prof. Tiow Seng Tan for their valuable suggestions. We extend our gratitude to the annotators for their hard work. We thank anonymous reviewers for their insightful feedback.

\bibliography{custom}
\clearpage
\appendix

\input{appendix/1_details_bafe}

\input{appendix/2_human_eval}

\input{appendix/3_long_fact_description_dataset}

\input{appendix/4_additional_autocorrelation_exps}

\input{appendix/5_counterfactual_examples}

\input{appendix/6_varying_length_responses_examples}

\input{appendix/7_prompts_empirical_studies}

\end{document}

%% file: tables/evaluation_method_comparison.tex
\begin{table}[t]
    \centering
    \resizebox{\columnwidth}{!}{%
    \begin{tabular}{cccc}
    \toprule
    \textbf{Method} & \makecell[c]{\textbf{Agreement} \\\textbf{w/ Humans}} & \makecell[c]{\textbf{Cost}\tablefootnote{Cost is calculated with \href{https://openai.com/api/pricing/}{OpenAI API Pricing} and \href{https://serper.dev}{Serper}}\\\textbf{(\$)}} & \makecell[c]{\textbf{Time} \\\textbf{(minute)}} \\
     \midrule
     \factscore & 69.97 & \textbf{0.021} & \textbf{0.67} \\
     \SAFE & 84.48 & 0.493 & 28.70\\
     \textbf{\BAFE(Ours)} & \textbf{89.31} & 0.067 & 7.17\\
     \bottomrule
    \end{tabular}%
    }
    \caption{Comparison of our method (\BAFE) with \SAFE~and \factscore. Agreement is measured against majority voting results from human annotators. Cost and time are calculated per response. Our method achieves the highest agreement with human annotations while being efficient.}
    \label{tab:method_comparison}
\end{table}

%% file: tables/counterfactual_analysis_results.tex
\begin{table}[t]
\small
    \centering
    \resizebox{\columnwidth}{!}{%
    \begin{tabular}{c|ccc}
    \toprule
    \textbf{Setting} & \makecell[c]{\textbf{First} \\\textbf{Sentence}} & \makecell[c]{\textbf{Second} \\\textbf{Sentence}} & \makecell[c]{\textbf{Subsequent} \\\textbf{Sentences}} \\
     \midrule
     Vanilla & 97.08 & 92.01 & 90.79 \\
     Flipped Factuality & 82.95 & 91.50 & 91.17\\
     \bottomrule
    \end{tabular}% 
    }
    \caption{Factual precision (\%) in vanilla and flipped factuality settings across three segments. ``Subsequent sentences'' refers to all sentences after the first in the response. Results show that factual errors in the first sentence do not propagate throughout the entire response.}
    \label{tab:counterfactual_analysis_results}
\end{table}

%% file: appendix/1_details_bafe.tex
\section{Details of \BAFE}
\label{appendix:detail_bafe}
This section provides additional details on the implementation and performance of \BAFE. We outline the implementation details of \BAFE~in Section~\ref{appendix:implementation_detail_bafe}. We present case studies comparing \BAFE~and \SAFE~in Section~\ref{appendix:case_comparison_safe}.

\subsection{Implementation Details of \BAFE}
\label{appendix:implementation_detail_bafe}
We first decompose long responses into atomic facts and then evaluate each atomic fact with bi-level verification. Detailed prompts for \BAFE~are available in our code.\footnote{Code is available at \url{https://github.com/XuZhao0/length-bias-factuality}.}

\paragraph{Response Decomposition.} We employ an LLM (\textit{gpt-3.5-turbo-instruct}) to split the responses into a series of atomic facts. We use few-shot prompting~\citep{10.5555/3495724.3495883} following \citep{min-etal-2023-factscore} and \citep{NEURIPS2024_937ae0e8}, and use greedy decoding for deterministic outputs. Each extracted atomic fact is then passed to the first-level verification stage.

\paragraph{First-Level Verification.} We follow the best practices in \factscore~and compare each atomic fact against a retrieved Wikipedia page. The verification step is conducted using an LLM, LLaMA~\citep{touvron2023llamaopenefficientfoundation}, which determines whether the fact is supported by Wikipedia. Given Wikipedia’s high reliability for general knowledge, we assume that facts supported at this level do not require further verification. Only atomic facts that are not supported by Wikipedia proceed to the second-level verification for broader fact-checking.

\paragraph{Second-Level Verification.} For atomic facts requiring second-level verification, we apply additional fact-checking using broader information sources. We first prompt an LLM (\textit{gpt-4o-mini-2024-07-18}) to revise each atomic fact to be self-contained. Using the same LLM, we then generate a single Google Search query for each atomic fact. Each query is issued to Google Search, retrieving the top 5 search results for comparison.\footnote{\SAFE~issues five queries per fact and considers the top 3 search results.} We include both ``title'' and ``snippet'' from search results for evaluation.\footnote{\SAFE~only uses ``snippets'' in the results, while we find that ``title'' will provide more information for factual judgment.} Additionally, we apply post-processing to reduce noise, such as removing misleading indicators like \textit{``Missing: <keywords>''}, which often lead to false positives. Given that Google Search aggregates broad and dynamic knowledge, if an atomic fact remains unsupported after both levels of verification, we classify it as unsupported.

\subsection{Case Study: Comparison of \BAFE~and \SAFE}
\label{appendix:case_comparison_safe}
In Section~\ref{sec:validate_bafe}, we demonstrate that \BAFE~achieves higher agreement with human annotations than \SAFE. To further illustrate this, we present two case studies in Table~\ref{tab:case_comparison_safe_1} and Table~\ref{tab:case_comparison_safe_2}, highlighting scenarios where our method makes correct judgments while \SAFE~fails.

\paragraph{Example in Table~\ref{tab:case_comparison_safe_1}.} \SAFE~relies solely on Google Search results, which consist of isolated snippets that may lack sufficient context. This can lead to incorrect factual judgments. For example, the statement \textit{``Throughout his career, Antonio Gasalla has appeared in numerous plays''} is supported by several paragraphs on Antonio Gasalla's Wikipedia page. However, since this information is not explicitly evident from Google Search snippets, \SAFE~incorrectly classifies it as unsupported. Additionally, \SAFE’s search results contain some distractions, such as \textit{``Missing: career plays.''}, further increasing the likelihood of errors.

\paragraph{Example in Table~\ref{tab:case_comparison_safe_2}.} Although \SAFE~issues five Google Search queries per atomic fact, its retrieved results often contain duplicate information, leading to redundancy and inefficiency. Furthermore, in cases where key factual details are absent, excessive search results introduce noise, causing the evaluator to make incorrect judgments. In this example, \SAFE~incorrectly assumes that \textit{``City of God''} was nominated for 50 Academy Awards due to repetitive and misleading search snippets. In contrast, \BAFE~issues only a single Google Search query, and carefully processes the top five results, leading to the correct judgment.

\begin{table*}[ht]
    \small
    \begin{tabular}{c|p{14cm}}
    \toprule
    \multicolumn{2}{l}{\textbf{\texttt{Statement}}: \texttt{Throughout his career, Antonio Gasalla has appeared in numerous plays.}}\\
    \midrule
    \midrule
    \makecell[c]{\textbf{Method}} & \makecell[c]{\textbf{Knowledge for Fact-Checking}} \\
     \midrule
     \SAFE & {\raggedright
\textit{\textbf{[First time search results]}} Antonio Gasalla (born March 9, 1941) is an Argentine actor, comedian, and theatre director. Antonio Gasalla. Born, (1941-03-09) March 9, 1941 (age 83). He is an actor and writer, known for El palacio de la risa (1992), Brother and Sister (2010) and La Argentina de Tato (1999). \textit{Missing: career plays performances.} 1. Waiting for the Hearse. 19851h 34m. 8.0 (7.6K); 2. The Truce. 19741h 48mNot Rated. 7.3 (735) ; 3. Hola Susana. 1987-20131997. 7.0 (34); 4. Brother and Sister. \textcolor{red}{\textit{Missing: career plays.}}\\
\textit{\textbf{[Second time search results]}} Antonio Gasalla (born March 9, 1941) is an Argentine actor, comedian, and theatre director. Antonio Gasalla. Born, (1941-03-09) March 9, 1941 (age 83). One of Argentina's most popular and longest-running TV shows on the air for over 50 years, where celebrities of all kinds sit down for lunch with Mirtha Legrand ... \textit{Missing: theater plays career.} He is an actor and writer, known for El palacio de la risa (1992), Brother and Sister (2010) and La Argentina de Tato (1999). \textcolor{red}{\textit{Missing: theater career}}\\
\textit{\textbf{...(Truncate for brevity)}}\\
        }  \\
    \midrule
     \makecell[c]{\BAFE\\(Ours)} & {\raggedright
Antonio Gasalla was born in Ramos Mejía, a western suburb of Buenos Aires, in 1941. He enrolled at the National Dramatic Arts Conservatory, and began his work in Buenos Aires' \textcolor{forestgreen}{vibrant theatre scene} in 1964 as an understudy, by which he befriended a colleague, Uruguayan émigré Carlos Perciavalle. He and Perciavalle starred in their production of \textcolor{forestgreen}{María Inés Quesada's Help Valentino! (1966)}, which they performed as a café-concert; this genre was popular in Argentina at the time, and the Gasalla-Perciavalle duo became among its best known exponents.\\
They accepted roles in film productions of Un viaje de locos (Madmen's Journey) and Clinica con musica \textcolor{forestgreen}{(Musical Clinic) in 1974}. Though known for their comedy roles, they were also cast in 1974 by Sergio Renán for \textcolor{forestgreen}{La tregua (The Truce)}, the first Argentine film nominated for an Oscar for Best Foreign Language Film. The duo parted ways subsequently, and Gasalla was cast in a comic role in \textcolor{forestgreen}{Tiro al aire (Shot in the Dark)}, a 1980 family film starring Héctor Alterio \\
\textit{\textbf{...(Truncate for brevity)}}\\
} \\
     \bottomrule
    \end{tabular}%
    \caption{A case study comparing \BAFE~and \SAFE. \SAFE~incorrectly classifies a supported fact as unsupported due to its reliance on isolated Google Search snippets. In contrast, \BAFE~leverages Wikipedia for more comprehensive verification, leading to a correct judgment. The key information is highlighted with \textcolor{forestgreen}{green} color.}
    \label{tab:case_comparison_safe_1}
\end{table*}

\begin{table*}[t]
    \small
    \begin{tabular}{c|p{14cm}}
    \toprule
    \multicolumn{2}{l}{\textbf{\texttt{Statement}}: \texttt{The film "City of God" was nominated for four Academy Awards.}}\\
    \midrule
    \midrule
    \makecell[c]{\textbf{Method}} & \makecell[c]{\textbf{Knowledge for Fact-Checking}} \\
     \midrule
     \makecell[c]{\SAFE} & {\raggedright
            \textit{\textbf{[First time search results]}} 75 wins \& 50 nominations. Academy Awards, USA. Fernando Meirelles at an event for The Constant Gardener (2005). 2004 Nominee Oscar. \textit{Missing: count | Show results with:count}. Full awards and nominations of City of God; Best Director…\\
            \textit{\textbf{[Second time search results]}} 75 wins \& 50 nominations. Academy Awards, USA. Fernando Meirelles at an event for The Constant Gardener (2005). 2004 Nominee Oscar. \textit{Missing: count | Show results with:count}. Full awards and nominations of City of God; Best Director…\\
            \textit{\textcolor{red}{Repeated several times...}}\\
            \textit{\textbf{[Fifth time search results]}} 75 wins \& 50 nominations. Academy Awards, USA. Fernando Meirelles at an event for The Constant Gardener (2005). 2004 Nominee Oscar. \textit{Missing: count | Show results with:count}. Full awards and nominations of City of God; Best Director…\\
        }  \\
    \midrule
  \makecell[c]{\BAFE\\(Ours)} & {\raggedright
Title: City of God (2002) \\
Awards - IMDb. 75 wins \& 50 nominations. Academy Awards, USA. Fernando Meirelles at an event for The Constant Gardener (2005). 2004 Nominee Oscar.\\
Title: Full awards and nominations of City of God - Filmaffinity.\\
Full awards and nominations of City of God nom. Best Director (Fernando Meirelles)  nom. Best Adapted Screenplay (Braulio Mantovani) nom. Best Film Editing ( ...\\
Title: City of God (2002 film) – Wikipedia.\\
City of God received widespread critical acclaim and garnered \textcolor{forestgreen}{four nominations at the 76th Academy Awards}; Best Cinematography (C\u00e9sar Charlone), Best Director ...\\
Title: City of God | Oscars Wiki - Fandom.\\
Nominations Best Adapted Screenplay 2014 Braulio Mantovani Best Cinematography 2014 Cesar Charlone  Best Director 2014 Fernando Meirelles  Best Film Editing 014 Daniel ...\\
Title: City Of God gets second wind after Oscar nominations - Screen Daily.\\
The surprise win of four key Oscar nominations by Brazil's gangland epic City Of God has prompted both Miramax in North America and Lumiere in ...
     } \\
     \bottomrule
    \end{tabular}%
    \caption{A case study comparing \BAFE~and \SAFE. \SAFE~issues multiple Google Search queries, resulting in redundant and noisy search results, which lead to an incorrect judgment. In contrast, \BAFE~uses a single query and carefully processes the top search results, enabling a correct factuality assessment. Key information is highlighted with \textcolor{forestgreen}{green} color.
    }
    \label{tab:case_comparison_safe_2}
\end{table*}

%% file: appendix/2_human_eval.tex
\definecolor{columbiablue}{rgb}{0.61, 0.87, 1.0}

\section{Details of Human Evaluation}
\label{appendix:human_eval}

\begin{figure*}[!t]
    \centering
    \includegraphics[width=\linewidth]{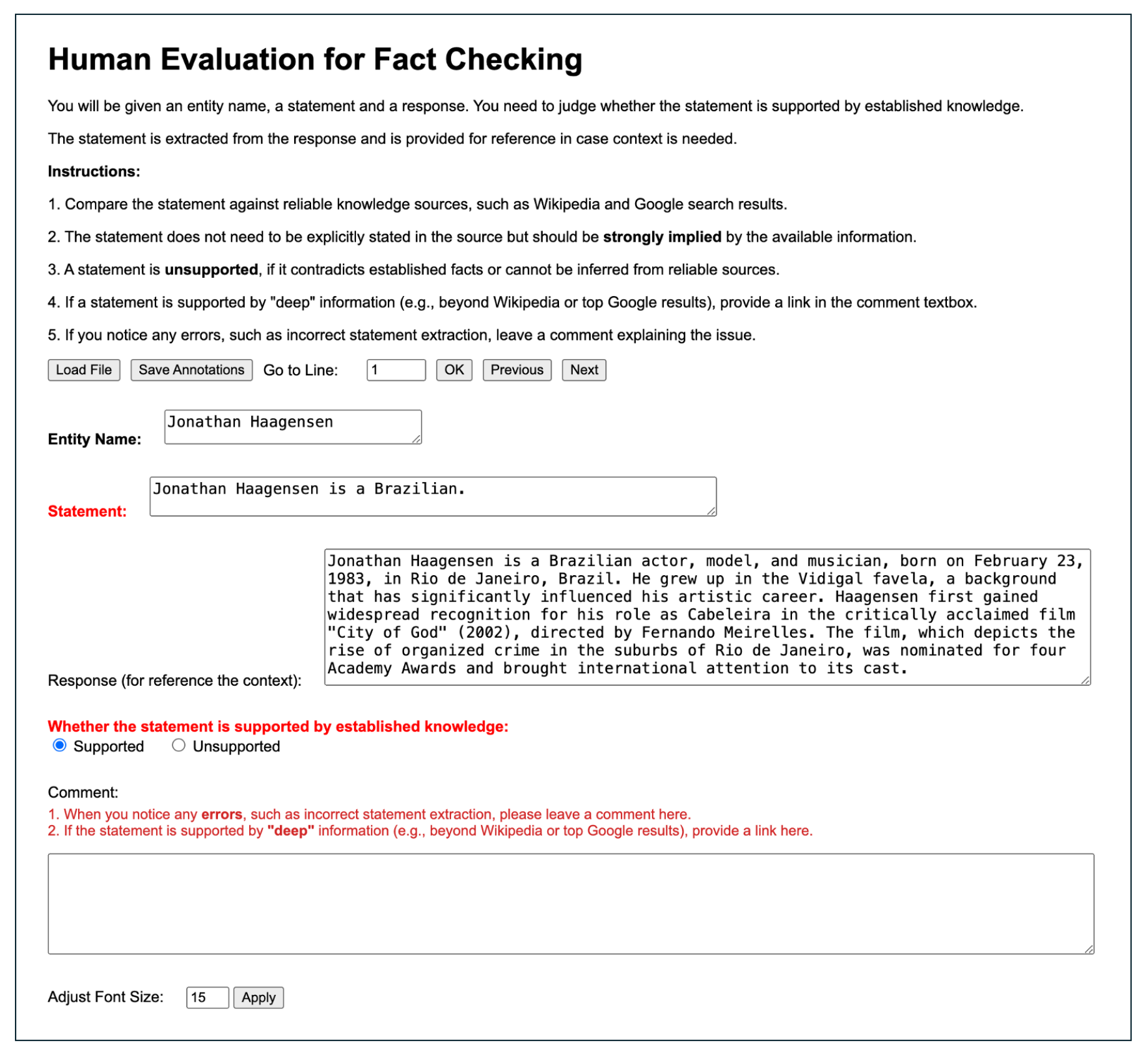}
    \caption{Human evaluation interface for fact-checking. Instructions are provided at the top of the interface. Annotators with full access to the Internet need to check whether the statement is supported by established knowledge.}
    \label{fig:human_eval_tool}
\end{figure*}

\paragraph{Human evaluation setup.} To verify the effectiveness of our method, \BAFE, we conducted a human evaluation. We recruited three university students as annotators and compensated them \$15 per hour. The annotation process took approximately 10 hours. All annotators were above 18 years of age and proficient in English. No other demographic and geographic restrictions were applied. No personally identifiable information was collected. All participants provided explicit consent for their annotations to be used in scientific publications. This study was approved by NUS Department Ethics Review Committee. Each annotator underwent a 40-minute training session, followed by a calibration phase where they labeled 50 sample data points. These labels were manually reviewed by us to clarify the annotation criteria. The human evaluation interface is shown in Figure~\ref{fig:human_eval_tool}. 

\paragraph{Evaluation data.} We randomly sample 18 responses generated by Llama-3.1-8B-Instruct, GPT-3.5-Turbo, and GPT-4o in the biography generation task, resulting in a total of 786 atomic facts for annotation. Among these, three human annotators fully agree on 685 cases, with a Fleiss $\kappa$ score at 0.7655 (substantial agreement). 

\paragraph{Analysis of annotation disagreements.} We manually review 50 cases in which the annotators give different judgments. Table~\ref{tab:human_disagreement_cases} categorizes these disagreement cases and provides representative examples. Most disagreements arise from differences in interpretation, varying levels of strictness in judging correctness, or ambiguity in statements. For example, the statement \textit{``City of Men is a TV series''} has varying interpretations, as both a TV series and a movie share the same name. This discrepancy results in different judgments. Another example is that the statement \textit{``Pharaoh Ramesses III established the stability of Egypt''} presents interpretational challenges. While some annotators consider it supported based on articles stating that Ramesses III defended Egypt against external threats, others argue that defense alone does not necessarily equate to establishing stability. Such differences in judgment contribute to annotation disagreements. Annotation errors account for only 24\% of the disagreements, and the overall disagreement rate remains low, indicating that the annotations are generally reliable.

\begin{table*}[t]
    \small
    \begin{tabular}{p{3.5cm}cp{10.5cm}}
    \toprule
    \makecell[c]{\textbf{Category}} & \textbf{\%} & \makecell[c]{\textbf{Examples}} \\
     \midrule
     {\raggedright Different interpretations of the factual information}  & 15 & {\raggedright
        \colorbox{columbiablue}{\texttt{Statement}} \texttt{``City of Men'' is a TV series.}\\
        \colorbox{melon}{\texttt{Comment}} The ``City of Men'' has varying interpretations, with some sources identifying it as a TV series and others as a movie. This discrepancy has resulted in differing judgments.
        }  \\
     \midrule
     {\raggedright Depends on the level of strictness in judging the correctness}  & 35 & 
     {\raggedright
        \colorbox{columbiablue}{\texttt{Statement}} \texttt{Pharaoh Ramesses III established the stability of Egypt.}\\
        \colorbox{melon}{\texttt{Comment}} Some annotators reckon the statement is true as Pharaoh Ramesses III indeed defended Egypt against external enemies, however, there were economic strains in which some annotators believe ``stability'' is overstated.
        }  \\
     \midrule
     {\raggedright Subjective statement with ambiguity}  & 26 & {\raggedright
        \colorbox{columbiablue}{\texttt{Statement}} \texttt{Jonathan Haagensen has showcased his appeal in the fashion industry.}\\
        \colorbox{melon}{\texttt{Comment}} On his Wikipedia page, it is stated that ``He has participated, as a model, in the Fashion Rio event, campaigning for Dolce and Gabbana.'' Some annotators inferred from this that he has appeal in the fashion industry.
        }  \\
     \midrule
     {\raggedright Mistakes in annotation}  & 24 & {\raggedright
        \colorbox{columbiablue}{\texttt{Statement}} \texttt{Antonio Gasalla has received a Konex Award.}\\
        \colorbox{melon}{\texttt{Comment}} Although Antonio Gasalla has received a Konex Award, this information is not explicitly stated on his Wikipedia page and requires a targeted search. This leads some annotators to mistakenly overlook the accolade.
        }  \\
     \bottomrule
    \end{tabular}%
    \caption{Categories and examples of human disagreement cases. \colorbox{columbiablue}{\texttt{Statement}} represents the statement being evaluated. \colorbox{melon}{\texttt{Comment}} indicates our comments.}
    \label{tab:human_disagreement_cases}
\end{table*}

%% file: appendix/3_long_fact_description_dataset.tex
\section{Long Fact Description Dataset}
\label{appendix:long_fact_dataset}

The long fact description dataset includes 140 non-person entities selected from LongFact-Concepts~\cite{wang-etal-2024-factuality}. It spans 4 broad categories: humanities, STEM, social science, and others. It is further subdivided into 26 topics, such as music and chemistry. Table~\ref{tab:longfact_details} presents the category, topics, entity examples, and the statistics.

\begin{table*}[t]
    \small
    \begin{tabular}{c|c|p{5cm}|p{5cm}}
    \toprule
    \textbf{Category} &  \textbf{Total Number} & \makecell[c]{\textbf{Topics}} & \makecell[c]{\textbf{Examples}}  \\
     \midrule
     Humanities & 37 & {\raggedright 20th-century-events, \\Architecture, \\International-law,  \\Jurisprudence,  \\Movies,  \\Music, \\World-religions}  & {\raggedright Hindenburg disaster,\\
Palace of Versailles, \\ Peace of Westphalia, \\ Loving v. Virginia, \\ The Big Short (film), \\ Blue Note Records, \\ Great Synagogue (Sydney)}  \\
\midrule
    STEM & 54 & {\raggedright Astronomy, \\ Biology, \\ Clinical-knowledge, \\Computer-science, \\Computer-security, \\ Machine-learning, \\ Mathematics, \\ Medicine, \\Physics, \\Virology} &{\raggedright Crab Nebula, \\ Eastern long-necked turtle, \\Karolinska Institute, \\Titan (supercomputer), \\Black Hat Briefings, \\GPT-3, \\Fields Medal, \\Johns Hopkins Hospital, \\Sardinia Radio Telescope, \\2009 swine flu pandemic}\\
\midrule
    Social Sciences & 36 &{\raggedright US-foreign-policy, \\Economics, \\Geography, \\History, \\Management, \\Prehistory, \\Sports} & {\raggedright Soweto uprising, \\International Monetary Fund, \\Antarctic Peninsula, \\Boston Tea Party, \\The Best Men Can Be, \\Chauvet Cave, \\Heisman Trophy}\\
\midrule
    Others & 13 & {\raggedright Accounting, \\Gaming } & {\raggedright Institute of Management, \\Accountants,\\Blizzard Entertainment}\\
     \bottomrule
    \end{tabular}%
    \caption{Distribution of entities in the \textit{long fact description} dataset by category, topic, and example. The dataset includes 140 non-person entities across 4 categories and 26 topics. }
    \label{tab:longfact_details}
\end{table*}

%% file: appendix/4_additional_autocorrelation_exps.tex
\section{Additional Experimental Results on Autocorrelation Analysis}
\label{appendix:error_propagation_more_models}

\begin{figure}[!t]
    \centering
    \includegraphics[width=\linewidth]{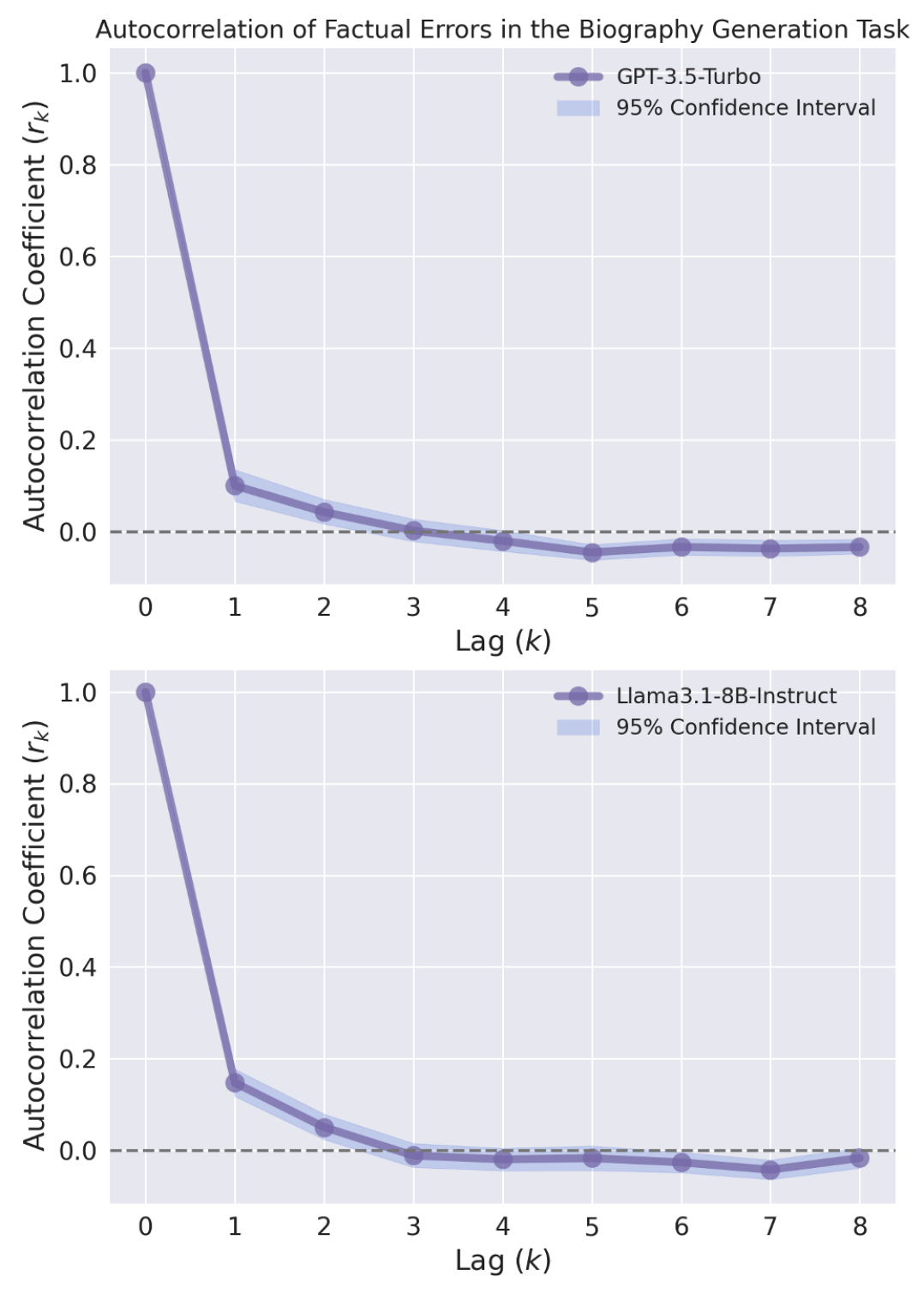}
    \caption{Autocorrelation coefficient at different lags on GPT-3.5-Turbo (\textbf{top}) and Llama-3.1-8B-Instruct (\textbf{bottom}) in the biography generation task. The 95\% confidence intervals are obtained via 2000 times bootstrap resampling. Only the coefficient at lag 1 is statistically higher than 0 for both models, suggesting weak short-term error propagation.}
    \label{fig:autocorrelation_more_models}
\end{figure}

To further examine the effect of error propagation, we extend our autocorrelation analysis to additional models and tasks. Specifically, we analyze \textit{GPT-3.5-Turbo} and \textit{Llama-3.1-8B-Instruct}\footnote{\href{https://huggingface.co/meta-llama/Llama-3.1-8B-Instruct}{https://huggingface.co/meta-llama/Llama-3.1-8B-Instruct}} in the biography generation task, and GPT-4o in the long fact description task.

\paragraph{Biography generation task.} Results for GPT-3.5-Turbo and Llama-3.1-8B-Instruct are shown in Figure~\ref{fig:autocorrelation_more_models}. Consistent with our earlier findings in Section~\ref{sec:autocorrelation_analysis}, only the coefficient at lag 1 is statistically higher than 0 for both models. The magnitude of the effect remains weak: Llama-3.1-8B-Instruct shows a coefficient of 0.18 at lag 1, and GPT-3.5-Turbo around 0.10. All subsequent lag values are close to zero, indicating that factual errors do not propagate beyond the immediate following fact.

\begin{figure}[!t]
    \centering
    \includegraphics[width=\linewidth]{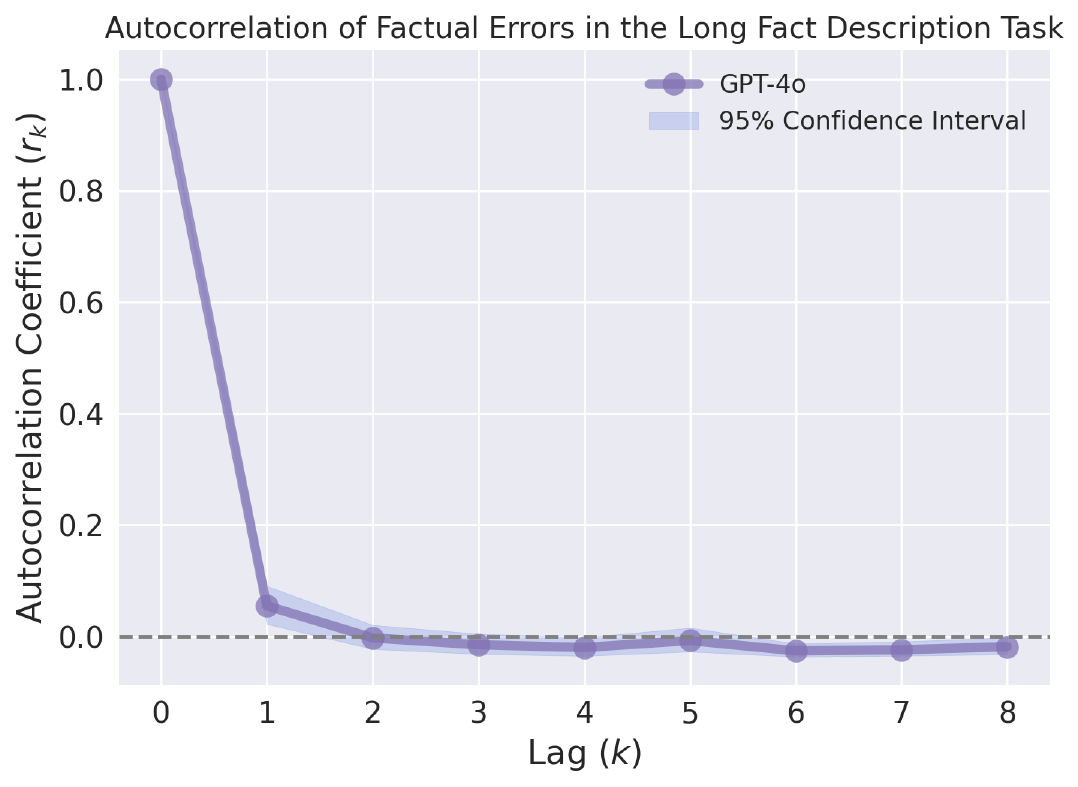}
    \caption{Autocorrelation coefficient on GPT-4o in the long fact description task. The 95\% confidence intervals are obtained via 2000 times bootstrap resampling.}
    \label{fig:autocorrelation_longfact}
\end{figure}

\paragraph{Long fact description task.} We also evaluate GPT-4o on the long fact description task. As shown in Figure~\ref{fig:autocorrelation_longfact}, the results show a similar trend: only the lag-1 coefficient is slightly above zero, indicating a weak short-term effect.

\paragraph{Conclusion.} Across different models (proprietary and open-source) and tasks, we observe a consistent pattern: error propagation exists but is weak and short-term effect. These results reinforce our conclusion that error propagation is not the primary cause of factual degradation in long-form generation.

%% file: appendix/5_counterfactual_examples.tex
\section{Counterfactual Analysis Examples}
\label{appendix:counterfactual_examples}

We present qualitative examples from counterfactual analysis to illustrate how early factual errors influence subsequent content. Specifically, we compare model responses generated from an original first sentence (vanilla setting) and a modified version (flipped factuality setting). As shown in Table~\ref{tab:counterfactual_response_examples}, flipping the factual correctness of the first sentence can lead to changes in response structure and content. The response in the flipped factuality setting is shorter and omits specific details, such as the birth date. It may help explain the slightly higher factual precision observed in the flipped factuality setting, as shown in Table~\ref{tab:counterfactual_analysis_results}.

\begin{table*}[t]
    \small
    \begin{tabular}{p{1.5cm}|p{13cm}}
    \toprule
    \makecell[c]{\textbf{Setting}} & \makecell[c]{\textbf{Responses}} \\
     \midrule
     \makecell[c]{Vanilla} & {\raggedright
\textit{Mike Trivisonno, often referred to as "Triv," was a prominent radio broadcaster based in \textcolor{forestgreen}{Cleveland}, Ohio.} \textit{Born on September 20, 1949}, in East Cleveland, Ohio, Trivisonno became a well-known voice in the region through his work on WTAM AM 1100, where he hosted "The Mike Trivisonno Show." His career in radio began in the late 1980s, and he quickly gained a reputation for his outspoken and often controversial style. Trivisonno's show covered a wide range of topics, including politics, sports, and local issues, and he was known for his ability to engage listeners with his candid opinions and sharp wit.\\
Trivisonno's influence extended beyond the airwaves, as he was actively involved in the community, often participating in charity events and fundraisers. He was particularly known for his support of local causes, including the Coats for Kids campaign, which provided winter clothing for children in need. Over the years, Trivisonno received numerous accolades for his work, including being named one of the top talk show hosts in the country by Talkers Magazine.\\
Despite his sometimes polarizing views, Trivisonno maintained a loyal following throughout his career, and his show remained a staple of Cleveland radio until his passing. Mike Trivisonno passed away on October 28, 2021, leaving behind a legacy as one of the most influential and recognizable figures in Cleveland broadcasting history. His impact on the community and the radio industry continues to be felt, and he is remembered for his passionate dedication to his craft and his city.
        }  \\
        \midrule
     \makecell[c]{Flipped \\Factuality} & {\raggedright
\textit{Mike Trivisonno, often referred to as "Triv," was a prominent radio broadcaster based in \textcolor{red}{Cincinnati}, Ohio.} He was best known for his long-running afternoon talk show on WTAM 1100 in Cleveland, Ohio, where he became a household name. Trivisonno's career in radio began in the late 1980s when he started as a frequent caller to the Pete Franklin show, earning the nickname "Mr. Know It All." His engaging personality and strong opinions quickly caught the attention of station executives, leading to his own show. Over the years, Trivisonno became known for his candid and often controversial discussions on a wide range of topics, including politics, sports, and local issues. His show was characterized by its interactive format, where listeners could call in and share their views, often leading to lively debates. Trivisonno's influence extended beyond radio as he became a well-known figure in the community, participating in various charitable events and fundraisers. Despite his sometimes polarizing views, he was respected for his dedication to his craft and his ability to connect with his audience. Trivisonno passed away on October 28, 2021, leaving behind a legacy as one of Cleveland's most iconic radio personalities.
    }  \\
     \bottomrule
    \end{tabular}%
    \caption{ Example responses in the vanilla setting and flipped factuality setting used in counterfactual analysis. The factual correctness of the first sentence is flipped from \textcolor{forestgreen}{\textit{Cleveland}} to \textcolor{red}{\textit{Cincinnati}}. The response in the flipped factuality setting omits some details, such as \textit{the birth date of Mike Trivisonno}, which is incorrect in the vanilla setting.
    }
    \label{tab:counterfactual_response_examples}
\end{table*}

%% file: appendix/6_varying_length_responses_examples.tex
\section{Example Responses with Varying Length}
\label{appendix:example_response_length}
In this section, we show example responses of varying length in Table~\ref{tab:appendix_response_length_1} and Table~\ref{tab:appendix_response_length_2}. These examples illustrate how increasing response length affects factual precision. As generation length increases, the model tends to include more speculative details, resulting in more factual errors. This observation further supports our finding that facts exhaustion is the primary cause of factual degradation in long-form text generation.

\begin{table*}[t]
    \small
    \begin{tabular}{c|p{13cm}}
    \toprule
    \makecell[c]{\textbf{Requested} \\\textbf{Length}} & \makecell[c]{\textbf{Responses}} \\
     \midrule
     100 & {\raggedright
            Nisar Ahmad Kakru is a distinguished Indian jurist who has made significant contributions to the legal field in India. \textcolor{red} {Born on October 25, 1949}, in the Baramulla district of Jammu and Kashmir, he pursued his education in law and embarked on a career that would see him rise to prominent judicial positions. Kakru served as the Chief Justice of the High Court of Andhra Pradesh from February 2010 to October 2011. Prior to this, he was a judge at the Jammu and Kashmir High Court, where he was appointed in 1997 and later became the acting Chief Justice. His tenure is noted for his commitment to justice and legal reform.
        }  \\
    \midrule
     200 & {\raggedright
             Nisar Ahmad Kakru is a distinguished Indian jurist known for his significant contributions to the judiciary in India.  \textcolor{red}{Born on October 25, 1949}, in the region of Jammu and Kashmir, Kakru pursued a career in law, which led him to become a prominent figure in the Indian legal system. He completed his law degree and began his legal practice, eventually gaining recognition for his expertise and dedication. Kakru's judicial career saw a steady rise as he was appointed as a judge of the Jammu and Kashmir High Court in 1997. \textcolor{red}{His tenure was marked by a commitment to justice and fairness, earning him respect among his peers} and the legal community. \textcolor{red}{In 2009, he was elevated to the position of Chief Justice of the Andhra Pradesh High Court}, where he served until his retirement in 2011. During his time as Chief Justice, Kakru was known for his efforts to improve the efficiency of the court and for his judicious handling of complex cases. Throughout his career, Nisar Ahmad Kakru has been recognized for his integrity, legal acumen, and dedication to upholding the rule of law. His contributions have left a lasting impact on the Indian judiciary, particularly in the regions he served.
     } \\
     \midrule
     300 & {\raggedright
            Nisar Ahmad Kakru is a distinguished Indian jurist known for his significant contributions to the judiciary in India.\textcolor{red}{Born on October 25, 1949}, in the region of Jammu and Kashmir, Kakru pursued a career in law, which led him to become a prominent figure in the Indian legal system. He completed his education in law and began his career as an advocate, practicing in various courts, including the Jammu and Kashmir High Court. Kakru's legal acumen and  \textcolor{red}{dedication to justice earned him a reputation as a fair and knowledgeable lawyer}, which eventually led to his appointment as a judge. He was appointed as a permanent judge of the Jammu and Kashmir High Court in 1997. \textcolor{red}{During his tenure, he was known for his judicious decisions} and his commitment to upholding the rule of law. \textcolor{red}{In 2009, Nisar Ahmad Kakru was appointed as the Chief Justice of the Andhra Pradesh High Court}, a position he held until his retirement in 2011.  His tenure as Chief Justice was marked by efforts to improve the efficiency of the court and to ensure that justice was accessible to all. He was known for his administrative skills and his ability to handle complex legal issues with clarity and fairness. Throughout his career, Justice Kakru has been recognized for his integrity, his deep understanding of the law, and his commitment to justice. His contributions to the Indian judiciary have been significant, and he is respected for his efforts to enhance the legal system in the regions he served. \textcolor{red}{After his retirement, Justice Kakru has continued to be involved in legal and social issues, contributing his expertise and experience to various causes.} His legacy in the Indian judiciary remains influential, and he is remembered as a judge who consistently worked towards the betterment of the legal system and society.
     }\\

     \bottomrule
    \end{tabular}%
    \caption{Examples of responses with varying lengths (100, 200, 300 words). Unsupported statements in the responses are highlighted in \textcolor{red}{red} color.}
    \label{tab:appendix_response_length_1}
\end{table*}

\begin{table*}[t]
    \small
    \begin{tabular}{c|p{13cm}}
    \toprule
    \makecell[c]{\textbf{Requested} \\\textbf{Length}} & \makecell[c]{\textbf{Responses}} \\
     \midrule
          400 & {\raggedright
            Nisar Ahmad Kakru is a distinguished Indian jurist known for his significant contributions to the judiciary in India, particularly in the state of Jammu and Kashmir. \textcolor{red}{Born on October 25, 1949}, in the picturesque region of Baramulla, Jammu and Kashmir, Kakru pursued a career in law that would see him rise to some of the highest judicial positions in the country. \textcolor{red}{Kakru completed his early education in his hometown before moving on to study law. He earned his law degree from the University of Kashmir, a prestigious institution in the region}, which laid the foundation for his future career in the legal field. After completing his education, Kakru began practicing law, where he quickly gained a reputation for his legal acumen and dedication to justice. \textcolor{red}{ His legal career took a significant turn when he was appointed as a judge of the Jammu and Kashmir High Court on March 7, 1994}. During his tenure as a judge, Kakru was known for his fair and balanced judgments, which earned him respect and recognition within the legal community. His work was characterized by a deep commitment to upholding the rule of law and \textcolor{red}{ensuring justice for all, regardless of their background}. \textcolor{red}{In 2009, Nisar Ahmad Kakru was appointed as the Chief Justice of the Andhra Pradesh High Court}, a position he held until his retirement in 2011. As Chief Justice, Kakru was instrumental in implementing several judicial reforms and \textcolor{red}{was known for his efforts to improve the efficiency and accessibility of the court system}. His leadership was marked by a focus on reducing case backlogs and enhancing the delivery of justice. Throughout his career, Kakru has been an advocate for judicial independence and  \textcolor{red}{has often spoken about the importance of maintaining the integrity of the judiciary.} His contributions to the legal field have been recognized by his peers and the public alike, and he remains a respected figure in Indian legal circles. After his retirement, Kakru continued to be involved in various legal and social initiatives, using his experience and knowledge to contribute to the betterment of society. His legacy is one of dedication to justice, legal excellence, and a commitment to the principles of fairness and equality. Nisar Ahmad Kakru's career serves as an inspiration to many aspiring lawyers and judges, highlighting the impact that a dedicated and principled individual can have on the legal system and society as a whole.
     }\\
     \midrule
     500 & {\raggedright
     Nisar Ahmad Kakru is a distinguished Indian jurist who has made significant contributions to the legal landscape of India, particularly in the state of Jammu and Kashmir. \textcolor{red}{Born on October 25, 1949}, in the picturesque region of Baramulla, Jammu and Kashmir, Kakru as had a long and illustrious career in the judiciary, marked by his commitment to justice and legal reform. \textcolor{red}{Kakru's early education was completed in his hometown}, after which he pursued higher studies in law. \textcolor{red}{He graduated with a degree in law from the University of Kashmir, which laid the foundation for his future career in the legal field.} His academic background and keen interest in law propelled him into the legal profession, where he began practicing as an advocate. Over the years, Kakru built a reputation as a competent and dedicated lawyer, which eventually led to his appointment as a judge. \textcolor{red}{His judicial career began when he was appointed as a permanent judge of the Jammu and Kashmir High Court on July 7, 2000}. During his tenure at the High Court, Kakru was known for his judicious approach and his ability to handle complex legal issues with clarity and fairness. One of the notable aspects of Kakru's career was his elevation to the position of Chief Justice of the Andhra Pradesh High Court. \textcolor{red}{He assumed this prestigious role on February 19, 2010, and served until October 24, 2011}. Kakru was instrumental in implementing several judicial reforms and was known for his efforts to improve the efficiency and accessibility of the judicial system.  \textcolor{red}{His leadership was characterized by a focus on reducing the backlog of cases} and ensuring timely justice for litigants. Throughout his career, Justice Kakru has been recognized for his integrity, impartiality, and dedication to upholding the rule of law. His judgments have often been cited for their depth of legal reasoning and adherence to constitutional principles.  \textcolor{red}{He has been a strong advocate for judicial independence and has consistently emphasized the importance of maintaining the judiciary's role as a guardian of the Constitution.} In addition to his judicial duties, Kakru has also been involved in various legal and educational initiatives. He has participated in numerous seminars and conferences, sharing his insights on legal issues and contributing to the discourse on judicial reforms.  \textcolor{red}{His commitment to legal education and mentorship has inspired many young lawyers and judges.} After retiring from active judicial service, Justice Kakru has continued to be a respected figure in the legal community. His contributions to the judiciary and his efforts to promote justice and legal reform have left a \textcolor{red}{ lasting impact on the legal system in India.} Nisar Ahmad Kakru's career is a testament to his unwavering dedication to the principles of justice and his significant role in shaping the judiciary in India. His legacy continues to influence the legal profession, and he remains a respected and admired figure among his peers and the broader legal community.
     }\\
      \bottomrule
    \end{tabular}%
    \caption{Examples of responses with varying lengths (400, 500 words). Unsupported statements in the responses are highlighted in \textcolor{red}{red} color.}
    \label{tab:appendix_response_length_2}
\end{table*}

%% file: appendix/7_prompts_empirical_studies.tex
\section{Prompts for Empirical Studies}
\label{appendix:example_prompts}

In this section, we provide prompts used in our empirical experiments. 
\begin{itemize}
    \item Table~\ref{tab:varying_length_prompt} and Table~\ref{tab:varying_length_prompt_longfact}: Prompt the models to generate responses of varying length, which is used to confirm the existence of length bias in Section~\ref{sec:length-bias}.
    \item Table~\ref{tab:vanilla_response_generation_prompt}: Prompt the model to generate a response at the model's default length, for investigating error propagation with autocorrelation analysis in Section~\ref{sec:autocorrelation_analysis}.
    \item  Table~\ref{tab:counterfactual_flip_fact}: Prompt the model to flip the factual correctness of the first sentence, serving as the start point for continued generation in Section~\ref{sec:counterfactual_analysis}. Note that we prompt the model three times to achieve the 82.95\% factual precision in Table~\ref{tab:counterfactual_analysis_results}.
    \item Table~\ref{tab:counterfactual_continue_generation}: Prompt used to continue generation from the first sentence. The generated responses are used in the counterfactual analysis in Section~\ref{sec:counterfactual_analysis}.
    \item Table~\ref{tab:context_length_prompt}: Prompt the model to generate both a context section and an evaluation section. It is used to analyze the effect of long context in Section~\ref{sec:effect_long_context}.
    \item Table~\ref{tab:fact_exhaustion_single_prompt} and Table~\ref{tab:fact_exhaustion_multiple_prompt}: Prompt used to generate responses in single-topic and multiple-topic settings, for the facts exhaustion experiments in Section~\ref{sec:effect_fact_exhaustion}.
\end{itemize}

\begin{table*}[t]
    \small
    \begin{tabular}{p{15cm}}
    \toprule
    \makecell[c]{\textbf{Prompt for Generating People Biographies with Varying Length}}  \\
     \midrule
     \ttfamily
     {\raggedright
\textbf{System Prompt:}\\
You are a helpful assistant. You will be given an entity name. You need to generate a bio for it. Here are the instructions:\\
1. The bio should be around \textit{<Length>} words.\\
2. Be sure to only include accurate, factual information in the response.\\
3. The bio should be comprehensive and detailed.\\
4. Do not include any controversial, disputable, or inaccurate factual claims in the response.\\
5. Return ONLY the bio, and nothing else.\\
\textbf{Instruction:}\\
Tell me a bio of \textit{<entity>}.
        }  \\
     \bottomrule
    \end{tabular}
    \caption{Prompt used for generating people biographies with varying requested output length. \texttt{<Length>} is set to \{100, 200,..., 500\} in our experiments. The generated responses are used to compute factual precision across different response lengths. The results are shown in Figure~\ref{fig:length_bias}.}
    \label{tab:varying_length_prompt}
\end{table*}

\begin{table*}[t]
    \small
    \begin{tabular}{p{15cm}}
    \toprule
    \makecell[c]{\textbf{Prompt for Generating Long Fact Descriptions with Varying Length}}  \\
     \midrule
     \ttfamily
     {\raggedright
\textbf{System Prompt:}\\
You are a helpful assistant. You will be given an entity related to <topic>. You need to provide a description of it. Here are the instructions:\\
1. The bio should be around \textit{<Length>} words.\\
2. Be sure to only include accurate, factual information in the response.\\
3. The bio should be comprehensive and detailed.\\
4. Do not include any controversial, disputable, or inaccurate factual claims in the response.\\
5. Return ONLY the bio, and nothing else.\\
6. Return the information in paragraph form using plain text, not in markdown or any other format.\\
\textbf{Instruction:}\\
Tell me about \textit{<entity>}.
        }  \\
     \bottomrule
    \end{tabular}
    \caption{Prompt used for generating long fact descriptions with varying requested output length. \texttt{<Length>} is set to \{100, 200,..., 500\} in our experiments. The generated responses are used to compute factual precision across different lengths. The results are shown in Figure~\ref{fig:length_bias_longfact}.}
    \label{tab:varying_length_prompt_longfact}
\end{table*}

\begin{table*}[t]
    \small
    \begin{tabular}{p{15cm}}
    \toprule
    \makecell[c]{\textbf{Prompt for Generating Biographies with the Model's Default Output Length}} \\
     \midrule
     \ttfamily
     {\raggedright
 \textbf{System Prompt:}\\
You are a helpful assistant. You will be given an entity name. You need to generate a bio for it. Here are the instructions:\\
1. Be sure to only include accurate, factual information in the response.\\
2. The bio should be comprehensive and detailed.\\
3. Do not include any controversial, disputable, or inaccurate factual claims in the response.\\
4. Return ONLY the bio, and nothing else.\\
\textbf{Instruction:}\\
Tell me a bio of \textit{<entity>}.
        }  \\

     \bottomrule
    \end{tabular}
    \caption{Prompt used for generating people biographies with the model's default output length. The generated responses are used for investigating the effect of error propagation with autocorrelation analysis in Section~\ref{sec:autocorrelation_analysis}.}
    \label{tab:vanilla_response_generation_prompt}
\end{table*}

\begin{table*}[t]
    \small
    \begin{tabular}{p{15cm}}
    \toprule
    \makecell[c]{\textbf{Prompt for Flipping the Factual Correctness}} \\
     \midrule
     \ttfamily
     {\raggedright
\textbf{System Prompt:}\\
You are a helpful assistant. You will be given a one-sentence bio of an entity. There are supported and unsupported facts in the bio. You need to convert one supported fact into an unsupported fact or generate new unsupported facts. And then you should give a new bio including the new unsupported facts. \\
The new bio should keep the syntax and structure of the original bio while introducing a small factual error. The new bio should still be one sentence.\\
Here are the guidelines for generating new unsupported facts:\\
1. Keep it plausible: The new unsupported facts should NOT alter the main point of the original bio. It should introduce small perturbations rather than major shifts in context.\\
2. The overall meaning should NOT change dramatically. **Small factual errors (e.g., places, dates, or minor career details) are suitable**.\\
3. You can generate unsupported facts by slightly altering the supported facts, referring to the original unsupported facts, or generating plausible but unsupported details, or in other ways.\\
4. Keep the provided unsupported facts in the new bio.\\
5. The inserted unsupported fact should relate to the broader biography and fit into the narrative.\\
You need to first give new unsupported facts. Then you need to give a new bio including the new unsupported facts. The new bio should match the format of the original bio as closely as possible.\\
The response format should be:\\
New unsupported facts: [new unsupported facts]\\
New bio: [new bio]\\
\textbf{Instruction:}\\
- Original bio: <the original first sentence>\\
- Supported fact: <all the supported atomic facts in the original first sentence>
        }  \\
     \bottomrule
    \end{tabular}
    \caption{Prompt used for flipping the factual correctness. The generated sentence is used in the flipped factuality setting in Section~\ref{sec:counterfactual_analysis}.}
    \label{tab:counterfactual_flip_fact}
\end{table*}

\begin{table*}[t]
    \small
    \begin{tabular}{p{15cm}}
    \toprule
    \makecell[c]{\textbf{Prompt for Continuing Generation from the First Sentence}} \\
     \midrule
     \ttfamily
     {\raggedright
\textbf{System Prompt:}\\
You are a helpful assistant. You will be given an entity name and the first sentence in the bio for it. You need to complete the given bio. Here are the instructions:\\
1. Be sure to only include accurate, factual information in the completed bio. \\
2. The completed bio should be comprehensive and detailed.\\
3. Do NOT change the given one-sentence bio. The completed bio should start with the given first sentence bio. \\
4. Return ONLY the completed bio, and nothing else.\\
\textbf{Instruction:}\\
Complete the following bio of <entity>.\\
The first sentence in the bio: <the first sentence>
        }  \\
     \bottomrule
    \end{tabular}
    \caption{Prompt used for continuing generation from the first sentence, which is either the original or a factually flipped version. The generated response is used in the counterfactual analysis in Section~\ref{sec:counterfactual_analysis}.}
    \label{tab:counterfactual_continue_generation}
\end{table*}

\begin{table*}[t]
    \small
    \begin{tabular}{p{15cm}}
    \toprule
    \makecell[c]{\textbf{Prompt for Generating Context and Evaluation Sections in Long-Context Experiments }}  \\
     \midrule
     \ttfamily
     {\raggedright
     \textbf{System Prompt:}\\
You are a helpful assistant. You will be given an entity name and two topics: \textit{"<Topic A>"} and \textit{"<Topic B>"}. You need to generate a bio for the entity that relates to the topics. Here are the instructions: \\
1. Firstly generate a bio relates to \textit{"<Topic A>"} with around \textit{<Context Length>} words.\\
2. Then generate a bio relates to \textit{"<Topic B>"} with around \textit{<Length>} words.\\
3. The response format should be like:\\
\#\#\# \textit{Topic A} \#\#\#\\
<Bio for Topic A>\\
\#\#\# \textit{Topic B} \#\#\#\\
<Bio for Topic B>\\
4. Be sure to only include accurate, factual information in the response.\\
5. The bio should be comprehensive and detailed.\\
6. Do not include any controversial, disputable, or inaccurate factual claims in the response.\\
7. Return ONLY the bio, and nothing else.\\
\textbf{Instruction:}\\
Tell me a bio of \textit{<entity>}.
        }  \\
     \bottomrule
    \end{tabular}
    \caption{Prompt used to investigate the effect of long context in Section~\ref{sec:effect_long_context}. \texttt{<Topic A>} and \texttt{<Context Length>} are used in the context section. \texttt{<Context Length>} is set to \{100, 200, ..., 600\} in our experiments. \texttt{<Topic B>} and \texttt{<Length>} are used in the evaluation section. \texttt{<Length>} is set to 200 in our experiments.}
    \label{tab:context_length_prompt}
\end{table*}

\begin{table*}[t]
    \small
    \begin{tabular}{p{15cm}}
    \toprule
    \makecell[c]{\textbf{Prompt for Generating Single-Topic Responses in Facts Exhaustion Experiments}}  \\
     \midrule
     \ttfamily
     {\raggedright
     \textbf{System Prompt:}\\
You are a helpful assistant. You will be given an entity name and one topic: \textit{"<Topic>"}. You need to generate a bio for the entity that relates to the topic. 
Here are the instructions: \\
1. Generate a bio relates to \textit{"<Topic>"} with around \textit{<Length>} words.\\
2. The response format should be like:\\
\#\#\# \textit{Topic} \#\#\#\\
<Bio for Topic>\\
3. Be sure to only include accurate, factual information in the response.\\
4. The bio should be comprehensive and detailed.\\
5. Do not include any controversial, disputable, or inaccurate factual claims in the response.\\
6. Return ONLY the bio, and nothing else.\\
\textbf{Instruction:}\\
Tell me a bio of \textit{<entity>}.
        }  \\
     \bottomrule
    \end{tabular}
    \caption{Prompt of single-topic setting in Section~\ref{sec:effect_fact_exhaustion}, for the investigation of facts exhaustion. \texttt{<Topic>} is set to either \textit{``Early life''}, \textit{``Personal life''} or \textit{``Career''}. \texttt{<Length>} is set to 400 in our experiments.}
    \label{tab:fact_exhaustion_single_prompt}
\end{table*}

\begin{table*}[t]
    \small
    \begin{tabular}{p{15cm}}
    \toprule
    \makecell[c]{\textbf{Prompt for Generating Multiple-Topic Responses in Facts Exhaustion Experiments}} \\
     \midrule
     \ttfamily
     {\raggedright
  \textbf{System Prompt:}\\
You are a helpful assistant. You will be given an entity name and two topics: \textit{"<Topic A>"} and \textit{"<Topic B>"}. You need to generate a bio for the entity that relates to the topics. 
Here are the instructions: \\
1. Firstly generate a bio relates to \textit{"<Topic A>"} with around \textit{<Length>} words.\\
2. Then generate a bio relates to \textit{"<Topic B>"} with around \textit{<Length>} words.\\
3. The response format should be like:\\
\#\#\# \textit{Topic A} \#\#\#\\
<Bio for Topic A>\\
\#\#\# \textit{Topic B} \#\#\#\\
<Bio for Topic B>\\
4. Be sure to only include accurate, factual information in the response.\\
5. The bio should be comprehensive and detailed.\\
6. Do not include any controversial, disputable, or inaccurate factual claims in the response.\\
7. Return ONLY the bio, and nothing else.\\
\textbf{Instruction:}\\
Tell me a bio of \textit{<entity>}.
        }  \\
     \bottomrule
    \end{tabular}
    \caption{Prompt of multiple-topic setting in Section~\ref{sec:effect_fact_exhaustion}, for the investigation of facts exhaustion. \texttt{<Topic A>} and \texttt{<Topic B>} are set to either \textit{``Early life''}, \textit{``Personal life''} or \textit{``Career''}. \texttt{<Length>} is set to 200 in the experiments.}
    \label{tab:fact_exhaustion_multiple_prompt}
\end{table*}